\documentclass[lettersize,journal]{IEEEtran}
\usepackage{amsmath,amsfonts}
\usepackage{algorithmic}
\usepackage{algorithm}
\usepackage{array}
\usepackage[caption=false,font=footnotesize,labelfont=rm,textfont=rm]{subfig}
\usepackage{textcomp}
\usepackage{stfloats}
\usepackage{url}
\usepackage{verbatim}
\usepackage{graphicx}
\usepackage{cite}
\usepackage{multirow}
\usepackage{etoolbox}
\usepackage{pifont}
\usepackage{xcolor}
\usepackage[colorlinks,
            linkcolor=black,
            anchorcolor=black,
            urlcolor=black,
            citecolor=black]{hyperref}
\usepackage{booktabs}
\usepackage{tabularx}  
\usepackage{caption}
\usepackage{cases}
\usepackage{float}

\definecolor{deepblue}{RGB}{0, 31, 175}

\begin{document}
\setlength{\skip\footins}{12pt} % 调整脚注和正文之间的距离

\title{Universal Trajectory Optimization Framework for Differential Drive Robot Class}

\vspace{-0.1cm}

\author{Mengke Zhang\textsuperscript{1}, Nanhe Chen\textsuperscript{1}, Hu Wang\textsuperscript{2}, Jianxiong Qiu\textsuperscript{2}, Zhichao Han\textsuperscript{1},\\ Qiuyu Ren\textsuperscript{1}, Chao Xu\textsuperscript{1}, Fei Gao\textsuperscript{1} and Yanjun Cao\textsuperscript{1}
        \thanks{$^*$This work was supported by National Natural Science Foundation of China under Grant 62103368. }
        % \thanks{$^1$The State Key Laboratory of Industrial Control Technology, College of Control Science and Engineering, Zhejiang University, Hangzhou 310027, China, and Huzhou Institute of Zhejiang University, Huzhou, 313000, China.}
        % \thanks{$^2$Zhejiang Zhongyan Industry Co. Ltd, Hangzhou, 310024, China.}
        \thanks{$^1$The State Key Laboratory of Industrial Control Technology, College of Control Science and Engineering, Zhejiang University, Hangzhou 310027, China. Huzhou Institute, Zhejiang University, and Huzhou Key Laboratory of Autonomous System, Huzhou 313000, China. }
        \thanks{$^2$China Tobacco Zhejiang Industrial Co., Ltd., Hangzhou 310024, China. }
        \thanks{Email:\tt\fontsize{7.8pt}{10pt}\selectfont\{mkzhang233,yanjunhi\}@zju.edu.cn}
        \vspace{-0.9cm}
}

% The paper headers
% \markboth{Journal of \LaTeX\ Class Files,~Vol.~14, No.~8, August~2021}%
% {Shell \MakeLowercase{\textit{et al.}}: A Sample Article Using IEEEtran.cls for IEEE Journals}

% \IEEEpubid{0000--0000/00\$00.00~\copyright~2021 IEEE}
% Remember, if you use this you must call \IEEEpubidadjcol in the second
% column for its text to clear the IEEEpubid mark.

\maketitle

\begin{abstract}
Differential drive robots are widely used in various scenarios thanks to their straightforward principle, from household service robots to disaster response field robots.
The nonholonomic dynamics and possible lateral slip of these robots lead to difficulty in getting feasible and high-quality trajectories. 
Although there are several types of driving mechanisms for real-world applications, they all share a similar driving principle, which involves controlling the relative motion of independently actuated tracks or wheels to achieve both linear and angular movement.
Therefore, a comprehensive trajectory optimization to compute trajectories efficiently for various kinds of differential drive robots is highly desirable.
In this paper, we propose a universal trajectory optimization framework, enabling the generation of high-quality trajectories within a restricted computational timeframe for these robots. 
We introduce a novel trajectory representation based on polynomial parameterization of motion states or their integrals, such as angular and linear velocities, which inherently matches the robots' motion to the control principle. 
The trajectory optimization problem is formulated to minimize computation complexity while prioritizing safety and operational efficiency.
We then build a full-stack autonomous planning and control system to demonstrate its feasibility and robustness. 
We conduct extensive simulations and real-world testing in crowded environments with three kinds of differential drive robots to validate the effectiveness of our approach.
\end{abstract}

\def\abstractname{Note to Practitioners}
\begin{abstract}
The differential drive robot, known for its simple mechanics and high maneuverability, is widely used in many applications. 
However, current methods have limitations in practice when high-performance motion is needed. 
Due to the state representation in Cartesian space, path planning makes it difficult to consider nonholonomic constraints directly. 
The existing trajectory optimization cannot effectively constrain the angular velocity and it is difficult to model forward and backward motion into a continuous trajectory. 
This paper provides a novel trajectory representation that inherently utilizes the motion performance of differential drive robots, which ensures its universality for different platforms, and reduces the time required to generate trajectories to ensure real-time performance.
Based on this, we propose a robust planning and control framework to achieve efficient navigation. 
We release the source code\footnote{\url{https://zju-fast-lab.github.io/DDR-opt/}}, facilitating expansion and deployment for practitioners.
We validate this framework through extensive experiments, demonstrating its capability to navigate challenging environments. 
\end{abstract}


\vspace{-0.5cm}

\section{Introduction}

\begin{figure}[t]
    \centering
    % \vspace{0.0cm}
    \setlength{\abovecaptionskip}{-2pt}
    \includegraphics[width=8.8cm]{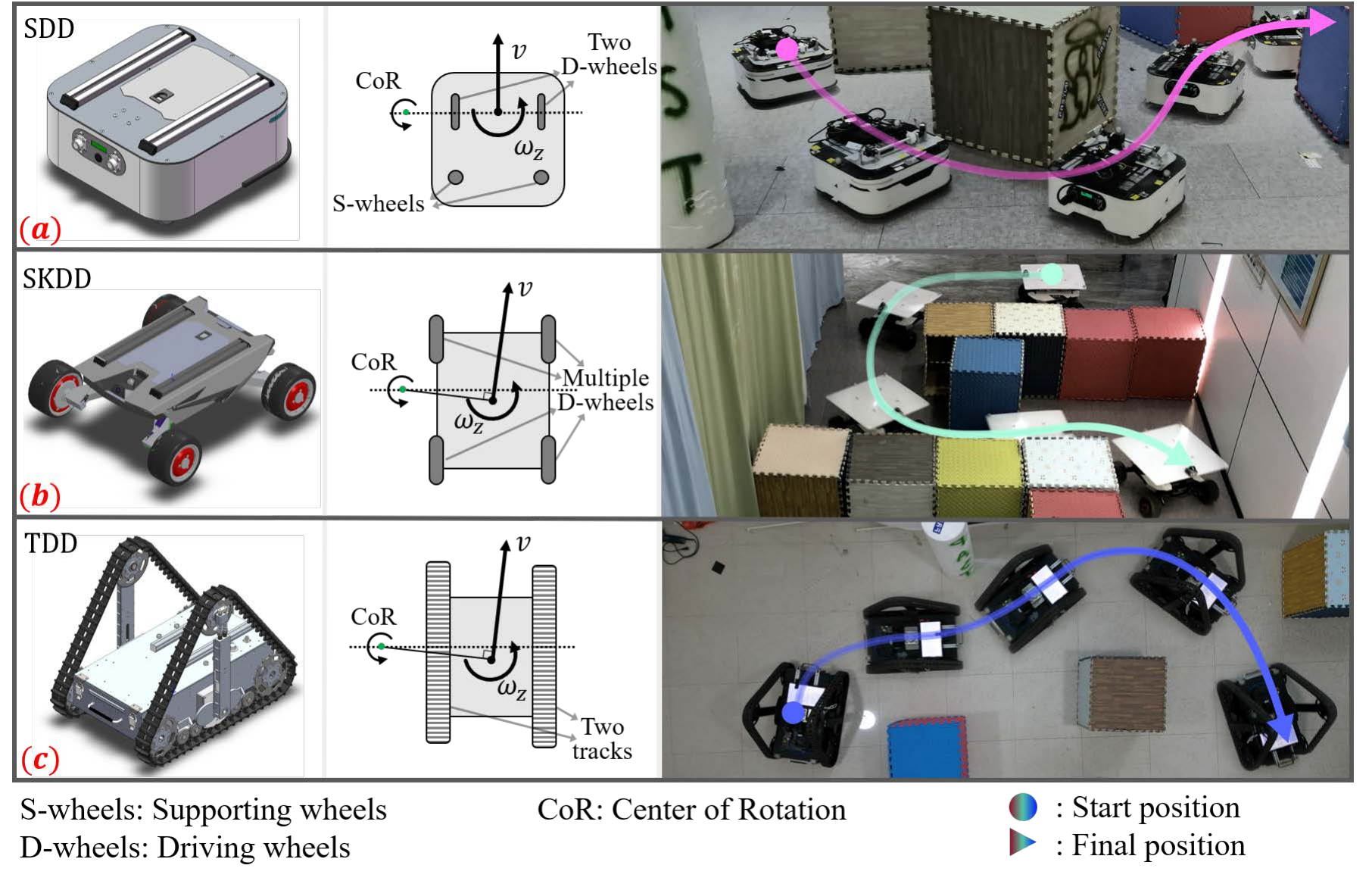}
    \caption{Various driving mechanisms of differential drive (DD) robots, along with their corresponding kinematic models and planning results.    
    (a) Two-wheeled standard differential drive (SDD) robot. (b) Skid-steering (SKDD) robot. (c) Tracked (TDD) robot. }
    \label{fig:head_figure}
    \vspace{-0.25cm}
\end{figure}

\begin{figure}[t]
    \centering
    \vspace{-0.4cm}
    \setlength{\abovecaptionskip}{1pt}
    \includegraphics[width=8.8cm]{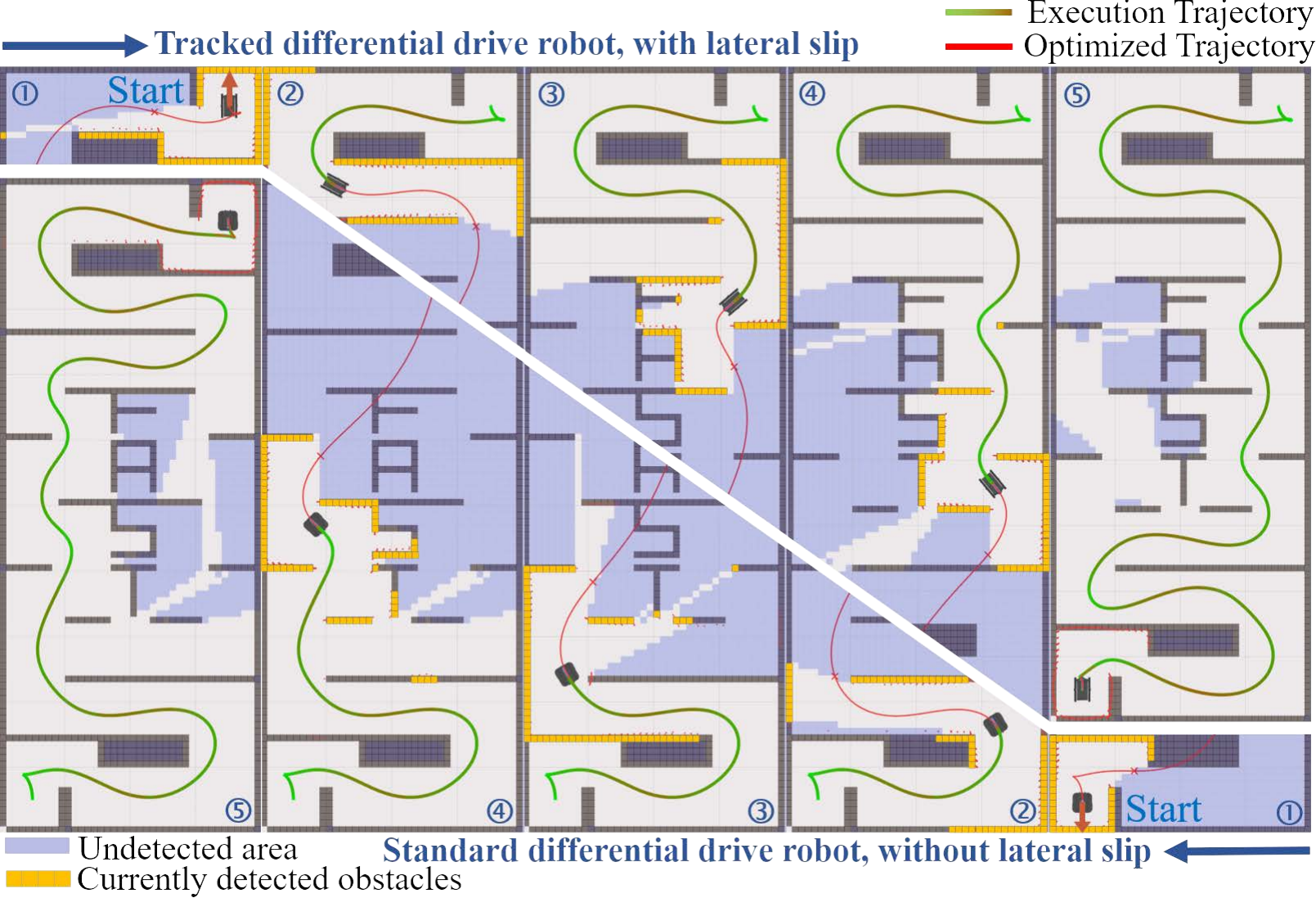}
    \caption{The optimized trajectory and the simulated execution results for two types of robots in narrow environments.  
    The robots should map online to perceive the environment and replan to avoid obstacles. 
    To verify the performance of the planner, a specifically designed map requires the robot to execute rotations or reversals at both the start and end points.
    In the upper right corner, from left to right, snapshots showcase the motion and mapping of the TDD robot, which moves with lateral slip.
    In the lower left corner, from right to left, is the SDD robot, which does not experience lateral slip. 
    }
    \label{fig:head_figure_trajectory}
    \vspace{-0.7cm}
\end{figure}

Differential drive (DD) has emerged as the predominant and widely adopted driving method for mobile robots due to its straightforward principle.
Various driving mechanisms are employed in real-world applications, including two-wheeled standard differential drive (SDD) robots, skid-steering differential drive (SKDD) robots, and tracked differential drive (TDD) robots, as illustrated in Fig. \ref{fig:head_figure}.
These robots share a common driving principle, where steering is achieved by controlling the speed difference between the wheels or tracks on either side, adhering to nonholonomic dynamics.
The nonlinearity of nonholonomic dynamics make it difficult to ensure the smoothness of states and control inputs in highly constrained environments. 
Ideally, DD allows for a minimal zero turning radius, making it particularly suitable for navigating crowded environments.
However, for robots equipped with tracks or multiple pairs of driving wheels, the interaction between the wheels or tracks and the ground causes the point with no normal velocity component to deviate from the geometric center of the robot, introducing lateral slip of the geometric center.
For trajectory generation, existing methods often face a trade-off: they either neglect lateral slip to simplify the planner, or compute short but precise trajectories to ensure computational efficiency.
Our objective is to develop a universal and efficient trajectory optimization framework for the DD robot class, considering nonholonomic constraints and potential lateral slip, thereby effectively harnessing the motion capabilities of these robots to generate executable trajectories.

Based on the aforementioned motivations, we abstract the desired attributes of the proposed algorithm into the following key criteria:  
(a) \textbf{Universality}: Despite the different driving mechanisms of DD platforms, they share a common driving principle.
The motion of these robots can generally be described by linear and angular velocities, which are related to the rotation speed of the driving wheels or tracks.
The trajectory should capture the motion characteristics of the DD robot class, incorporating concepts such as instantaneous centers of rotation (ICRs) \cite{martinez2005approximating} to accurately model the kinematics.
The corresponding trajectory optimization method is also required to get an executable optimal trajectory.
A universal method that meets these requirements supports further algorithm development, facilitating a more maintainable planning framework and promoting easier collaboration across multiple platforms.
(b) \textbf{Trajectory Quality}: The robot's trajectory should be feasible and smooth for the controller to decrease trajectory tracking errors and reduce energy consumption. 
Movement in different directions should be modeled into a single trajectory, including forward and backward movements. 
The kinematics of robots need to be easily modeled by the trajectory.
The spatiotemporal optimization is also necessary.  
(c) \textbf{Computational Efficiency}: Navigating in complex environments usually requires frequent replanning and therefore the time to replan is important for system performance. 
The algorithm should generate feasible trajectories within a limited time. 
This flexibility not only guarantees the continuity of planning but also allows the robot to quickly respond to updated scenarios, thereby enhancing its adaptability in challenging environments. 

Although various planning methods have been studied, to the best of our knowledge, no work has been found that fully meets the aforementioned requirements. 
Traditional search-based methods, despite their universality with various kinematic models, struggle to generate high-quality trajectories within limited discrete samples. 
Strategies that utilize parameterization of the robot's position, such as the Differential Flatness-based (DF) approach, offer improved computational efficiency. 
However, they may introduce singularities \cite{zhang2023trajectory}, resulting in nonlinearity in the state expressions, particularly at low velocities.
As demonstrated in \ref{subsec:replan}, these singularities can adversely affect the planner's performance.
Methods based on Optimal Control Problem (OCP) discretize the trajectory and optimize control inputs, which require solving larger-scale nonlinear programming problems. 
The presence of nonholonomic constraints and lateral slip makes it challenging to model position trajectories, and the discretization of control inputs struggles to ensure smoothness and computational efficiency.

In this work, we propose a novel trajectory representation method based on polynomial parameterization of \textbf{M}otion \textbf{S}tates, abbreviated as MS trajectory hereinafter. 
We directly parameterize the robot's motion states or their integrals, such as linear and angular velocity, ensuring that the trajectories naturally satisfy the robot's nonholonomic constraints while effectively modeling lateral slip. 
Optimizing the motion states guarantees that the kinematic constraints are linear combinations of the optimization variables, ensuring optimality. 
At the same time, this parameterization leads to the control inputs being calculated by smoother and less complex polynomials compared to OCP, thereby reducing computational complexity. 
The application of motion states-based numerical integration allows analytically transforming the trajectory into positions, enabling constraints of the Cartesian space. 
The computational efficiency and universal trajectory representation provide feasible trajectories applicable to DD platforms with similar characteristics.

Building upon the trajectory representation, we have developed a universal trajectory planning system for DD robots that efficiently generates high-quality trajectories. 
To ensure robustness, we employ trajectory preprocessing to avoid topological changes caused by the initial value and front-end path errors. 
Considering the uncertainty in kinematic parameters when using wheel speed control, we use the Extended Kalman Filter (EKF) to estimate the kinematic parameters and design the Nonlinear Model Predictive Control (NMPC) to track the desired trajectory based on this estimation.
The optimized trajectory and the simulated execution results in narrow environments are shown in Fig.\ref{fig:head_figure_trajectory}. 

The main contributions of this paper are as follows:

1. A universal MS trajectory representation method to describe the motion of the DD robot class from motion states, which can describe nonholonomic dynamics and lateral slip. 

2. An efficient MS trajectory-based optimization method that can quickly compute smooth trajectories while satisfying the robot's kinematic and safety constraints.

3. A robust trajectory planning and control framework applicable to various DD platforms. 

4. Extensive simulations and real-world experiments validate the effectiveness of the proposed method, which will be released as an open-source package.

% The rest of this paper is organized as follows. 
% Sec.\ref{sec:related_work} reviews the related work in the field.  
% Sec.\ref{sec:Notations} introduces the commonly used symbols. 
% Sec.\ref{sec:Traj_rep} introduces the MS trajectory and discusses the usage of numerical integration to get positions. 
% Sec.\ref{section:Optimization_Problem} introduces the optimization problem based on the new trajectory representation. 
% Sec.\ref{sec:constraints} focuses on the specific forms of constraints for DD robots. 
% Sec.\ref{sec:navi_sys} introduces the whole system for robust planning and control.
% Sec.\ref{sec:sim_exp} and Sec.\ref{sec:real_exp} validate the proposed method through simulation and real-world experiments, respectively.
% Sec.\ref{sec:conclusion} concludes the paper.

\section{Related Work}\label{sec:related_work}

Motion planning of DD robots has been extensively studied. 
Heuristic function-based methods \cite{dolgov2010path, harabor2011online, gupta2017dynamically, zhou2020trajectory, 9627934} and sampling-based methods \cite{webb2013kinodynamic, burget2016bi, wang2020neural, wang2020eb, 9970002} are widely used in robot motion planning due to their easy implementation to handle user-defined constraints. 
These methods usually discretize the environment and generate a feasible path connecting the starting and final positions through graph search or sampling. 
However, they could be computationally expensive when taking kinematics or dynamics into account. 
Some methods \cite{9970002,9638379} have considered the kinematics and can ensure motion continuity. 
However, they still require a strict and difficult-to-achieve balance between computational cost and path quality.
Methods such as DWA \cite{fox1997dynamic, yang2022automatic} and LQR \cite{cornejo2018trajectory, ni2022path} can be used for robot control and are widely used thanks to the kinematic capabilities of DD robots, but their optimality cannot be guaranteed.

For relatively simple SDD robots, studies \cite{heiden2018gradient, jian2022parametric} optimize paths by interpolating and optimizing key waypoints. 
Wang et al.\cite{wang2019ogbps} get optimal paths by minimizing the path length and maximum curvature without taking temporal information into consideration. 
Although interpolation methods can be applied to give temporal information, the spatial and temporal optimality of the trajectories cannot be guaranteed. 
Kurenkov et al.\cite{kurenkov2022nfomp} use Chomp\cite{ratliff2009chomp} to plan the position and orientation of the robot and constrain the nonholonomic dynamics using the Lagrangian method, but the presence of equation constraints makes it computationally inefficient.

Recently, polynomial-based trajectory optimization methods \cite{zhou2019robust, zhou2020ego,wang2022geometrically} have been applied to aerial robot motion planning. 
These methods are typically employed for quadrotors without nonholonomic dynamic constraints.
They can also be applied to trajectory optimization for ground robots with the differential flatness property \cite{han2023efficient, zhang2023trajectory}, achieving optimal trajectories with high efficiency.
However, differential flatness introduces singularities due to nonholonomic dynamics, resulting in poor optimization performance at low speeds and relying on the front-end methods like hybrid A*\cite{dolgov2010path} to provide an initial path for choosing forward or reverse motion. 
Xu et al.\cite{xu2023efficient} additionally plan the orientation angle and use the augmented Lagrangian method to deal with nonholonomic dynamics, but multiple iterations are required to satisfy the equation constraints, which is less efficient. 
In addition, differential flatness cannot model the lateral slip, as the orientation is defined as the velocity direction of the center of mass. 

Due to the unavoidable lateral slip inherent in the mechanical principles of SKDD and TDD robots, they usually require more accurate modeling approaches, including both dynamic models \cite{yu2010analysis} and kinematic models \cite{martinez2005approximating, fink2014experimental, zhao2019kinematics, wang2015analysis}.
Many methods \cite{zhao2019kinematics, yi2009kinematic, pentzer2014model, qin2021improved} apply kinematic models to estimate parameters to reduce tracking errors, which maintaining good computational efficiency. 
Some methods \cite{pace2017experimental} incorporate the kinematic parameters of robots into motion planning, but usually need to oversimplify the kinematic model. 
Methods based on optimal control \cite{huskic2019high, rigatos2021nonlinear} and MPC variants \cite{dorbetkhany2022spatial, prado2020tube} are also used to compute the trajectory. 
However, these methods can only compute trajectories in a relatively short horizon for real-time performance and cannot guarantee global optimality.
Rosmann et al.\cite{rosmann2012trajectory, rosmann2013efficient} take into account the time and velocity of lateral slip and present the Timed Elastic Band (TEB) planner, based on a hyper-graph and implemented with G\textsuperscript{2}o\cite{kummerle2011g}.
However, it may take a long time to generate an optimal trajectory, and considering the velocity of lateral slip individually for DD robots is unreasonable.

Our method proposes a novel \textit{motion state}-based trajectory representation and trajectory optimization algorithm taking advantage of polynomial-based trajectory optimization and state propagation. 
By using fewer variables to represent long trajectories, we balance computational cost and the ability to characterize complex motion. 
The \textit{motion state}-based method allows the trajectory to satisfy the kinematic model inherently. 
We introduce online parameter estimation and NMPC controllers based on this approach, which leads to improved tracking performance of the optimized trajectories.

% \section{Notations}\label{sec:Notations}

% Table.\ref{tab:Notations} shows the notations used across different sections of this paper, including variables and concepts related to the robot's state, superscripts and subscripts for clarity, and trajectory representation.
% For brevity and to maintain a fluent narrative, these notations will not be reiterated in subsequent sections. 
% Only variables that are used once or within a single section are explained when introduced.
\section{Trajectory Representation}\label{sec:Traj_rep}
In this section, we focus on a novel trajectory representation method based on the polynomial parameterization of motion states.
Table.\ref{tab:NotationsTra} shows the notations used in this section, including variables and concepts related to the robot's state, superscripts and subscripts for clarity, and trajectory representation. 

\begin{table*}[t]
    \caption{Notations of Trajectory}
    \vspace{-3pt}
    \centering
    \label{tab:NotationsTra}
    \fontsize{9}{11}\selectfont % 设置字体大小为12pt，行距为14pt
    % \begin{tabular}{@{}c p{6.5cm}@{}}
    \begin{tabularx}{\textwidth}{@{}c X@{}}  
    \toprule
    $\textbf p=[x,y,\theta]^T$  & The position and orientation of the body frame.   \\
    $v_x, v_y, \omega$ & The linear velocity along the x-axis and y-axis, and angular velocity in the body frame. \\
    $V_r, V_l$ & The speeds of the right and left wheels or tracks, respectively. \\
    $x_{Iv}, y_{Iv}, y_{Il}, y_{Ir}$ & Instantaneous Centers of Rotation (ICRs) \cite{martinez2005approximating}. \\
    \midrule 
    $(\cdot)(t)$ & The value of the trajectory at time $t$. \\
    $\tilde {(\cdot)}_f$ & The final position calculated by integration. \\
    $\bar {(\cdot)}_i$ & The position increment of the $i$th segment of the trajectory. \\
    $\hat {(\cdot)}^{j,l}_i$ & The integrand function value at the $l$th sampling point within the $j$th sampling interval of the $i$th segment. \\
    $T_i^{j,l}$ & The timestamp of the sampling point $\hat {(\cdot)}^{j,l}_i$, where $T_i^{j,l} = \frac{2(j-1)+l}{2n}T_i$. \\
    ${(\cdot)}^{i,j}$ & The corresponding value after the $j$th sampling interval of the $i$th segment. \\
    ${(\cdot)}_0, {(\cdot)}_f$ & The initial and final conditions of the trajectory. \\
    \midrule     
    $s$ & The accumulated forward arc length along the x-axis of the robot's body frame.  \\
    $M$, $n$ & The number of polynomial segments of the trajectory and the number of sampling intervals for each segment. \\
    $\boldsymbol c_{\theta,i}, \boldsymbol c_{s,i}, \boldsymbol c_i$ & The $i$th segment's coefficient vectors of $\theta$, $s$ and both,  respectively, where $\boldsymbol c_i=[\boldsymbol c_{\theta,i},\boldsymbol c_{s,i}]$.  \\
    $\boldsymbol\beta(t)$ & The basis of the polynomial. \\
    $T_i$& The duration of the $i$th segment. \\
    \bottomrule
    \end{tabularx}  
    \vspace{-18pt}
\end{table*}

\subsection{Kinematic Model of DD Robots}\label{sec:Kinematic_Model}

% Starting with the SDD robots, we define the states as $\textbf p=[x,y,\theta]^T$. 
% The origin of the body frame is situated at the midpoint between the two driving wheels, with the x-axis extending in the direction of the robot. 
% The motion of the base is controlled by the wheel speed input $\textbf u=[V_r,V_l]^T\in\mathbb R^2$.
% Thus, the kinematics of the base can be expressed as: 
% \begin{align}
%     \begin{bmatrix} v_x \\ \omega \end{bmatrix} & = \begin{bmatrix} \frac{1}{2} & \frac{1}{2} \\ \frac{1}{d_{wb}} & -\frac{1}{d_{wb}} \end{bmatrix} \begin{bmatrix} V_r \\ V_l  \end{bmatrix}, \label{ali:vxomegaz}
% \end{align}

% where $d_{wb}$ is the wheelbase of the driving wheels. 

\begin{figure}[b]
    \centering
    \vspace{-0.75cm}
    \setlength{\abovecaptionskip}{2pt}
    \includegraphics[width=8.8cm]{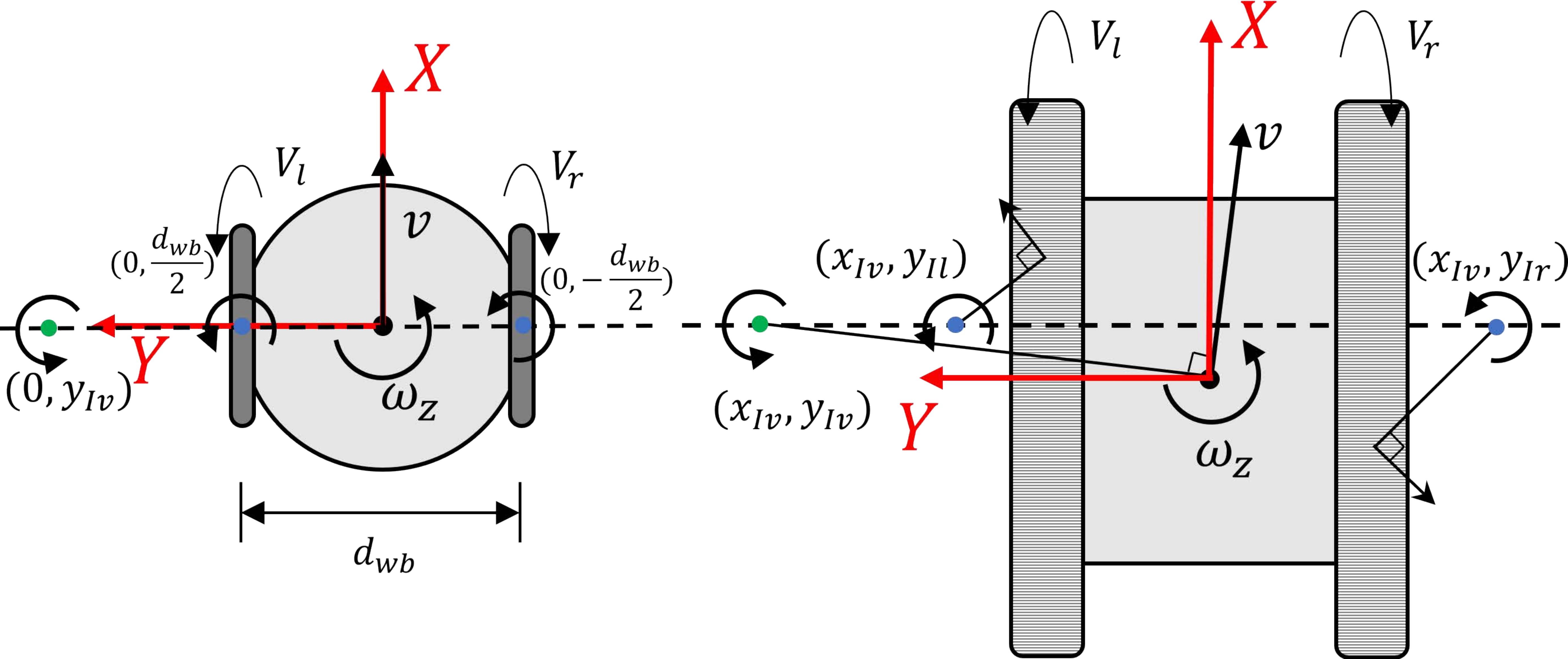}
    \caption{Kinematic models of SDD (left) and TDD (right) robots. 
    The green point represents the ICR of the robot's body, and the blue points represent the ICRs at the contact points between the driving wheels and the ground. 
    For tracked robots, slipping causes the ICRs of the tracks to misalign with their ground projections. 
    $(x_{Iv}, y_{Il}), (x_{Iv}, y_{Ir}), (x_{Iv}, y_{Iv})$ represent the positions of the ICRs of the left track, right track, and the body, respectively, in the robot's body coordinate frame.
    }
    \label{fig:tracked_model}
    % \vspace{-0.4cm}
\end{figure}

We use the instantaneous centers of rotation (ICRs) \cite{martinez2005approximating} to estimate the motion of the DD robot, as illustrated in Fig.\ref{fig:tracked_model}. 
By employing ICRs, we can conveniently define the origin of the robot's body coordinate frame as the geometric center of the robot, with its x-axis aligned with the robot's orientation. 
The robot's instantaneous translational and rotational velocities can be expressed as follows:  
\begin{align}
    \omega & = \frac{V_r - V_l}{y_{Il} - y_{Ir}}, \label{ali:omegaz} \\  
    v_x & = \frac{V_r + V_l}{2} - \frac{V_r - V_l}{y_{Il} - y_{Ir}} \left(\frac{y_{Il} + y_{Ir}}{2}\right), \label{ali:vx} \\  
    v_y & = -\frac{V_r - V_l}{y_{Il} - y_{Ir}} x_{Iv} = -\omega x_{Iv}. \label{ali:vy}  
\end{align}

For SDD robots, the ICRs of the wheels are located at their respective contact points with the ground, with $y_{Il} = -y_{Ir} = d_{wb}/2$, where \(d_{wb}\) is the wheelbase of the driving wheels, and no sideslip occurs ($x_{Iv} = 0$). 
The body coordinate frame's origin is positioned at the midpoint between the two driving wheels.  
To maintain universality, we utilize Eqs.(\ref{ali:omegaz})--(\ref{ali:vy}) for trajectory computation across different driving mechanisms. 

\subsection{Motion State Trajectory}
Unlike the classical trajectory representation in Cartesian coordinates, we express the trajectory directly in the motion states domain.
This trajectory consists of the orientation $\theta$ and the forward arc length $s$ as polynomials of time:
\begin{equation}
    \begin{aligned}[c]
        \theta_i(t)=\boldsymbol \beta^T(t) \boldsymbol c_{\theta,i}, \\
        s_i(t)=\boldsymbol \beta^T(t) \boldsymbol c_{s,i}, \label{ali:theta_s}
    \end{aligned}
\end{equation}
where $\boldsymbol\beta(t)=[1,t,t^2,\ldots,t^{2h-1}]^T$ is the natural basis. 
The derivatives of Eq.(\ref{ali:theta_s}) are the motion states, such as linear velocity $v_x=\dot s$ and angular velocity $\omega = \dot\theta$. 
To distinguish it from the widely used trajectory defined by positions $x$ and $y$, we refer to this trajectory as the \textbf{M}otion \textbf{S}tate (MS) trajectory.
We divide the trajectory into $M$ segments, each of which is represented as a 2-dimensional polynomial of order $2h-1$. 

Given that the final state is defined by the position in the Cartesian frame, we need to transform the position from MS trajectory to Cartesian coordinates by integration. 
The x-coordinate of the robot at any time $t$ can be calculated as:
\begin{equation}
    \begin{aligned}[c]
        x(t) & = \int_0^t[v_x(\tau)\cos\theta(\tau) - v_y(\tau)\sin\theta(\tau)]d\tau + x_0 \\
             & = \int_0^t [\dot s(\tau)\cos\theta(\tau) + x_{Iv}\dot\theta(\tau)\sin\theta(\tau)] d\tau + x_0.     \label{ali:intergalx}
    \end{aligned}
\end{equation}
Similarly, the y-coordinate can be calculated as: 
\begin{equation}
    \begin{aligned}[c]
        y(t) = \int_0^t [\dot s(\tau)\sin\theta(\tau) - x_{Iv}\dot\theta(\tau)\cos\theta(\tau)] d\tau + y_0,
        \label{ali:intergaly}
    \end{aligned}
\end{equation}
where $(x_0, y_0)$ is the starting position. 
However, obtaining an analytic solution for Eqs.(\ref{ali:intergalx}-\ref{ali:intergaly}) directly is difficult. 
Therefore, we choose a numerical integration method to approximate it. 
Specifically, we employ Simpson's rule, which allows us to represent the integral as a weighted sum of sampling points: 
\begin{align}
    \int_a^bf(x)dx\approx\dfrac{b-a}{6}[f(a)+4f(\dfrac{a+b}{2})+f(b)], 
    \label{ali:Simpson}
\end{align}
where $f(x)$ is the integrand function and $[a,b]$ is the integration interval. 

Given that gradients are required during optimization, we need to consider both the accuracy of the integral and the computational complexity for the MS trajectory form.
Simpson's rule inherently has low computational complexity by using lower-order polynomials and a small number of sampling points for the approximation.
We divide each segment into $n$ intervals, and each interval performs the approximation of Eq.(\ref{ali:Simpson}). 
Taking the $x$-axis as an example, the integral approximation for the entire trajectory can be expressed as:
\begin{align}
    &\tilde x_f = \sum\limits_{i=1}^M \overline x_i(\boldsymbol c_i, T_i)+x_0, \nonumber\\
    &\overline x_i(\boldsymbol c_i, T_i) =\dfrac{T_i}{6n}\sum\limits_{j=1}^{n}(\hat x^{j,0}_i+4\hat x^{j,1}_i+\hat x^{j,2}_i),  \label{ali:AppSim}\\
    &\hat x^{j,l}_i=\dot s_m(T_i^{j,l})\cos(\theta_i(T_i^{j,l})) + x_{Iv}\dot\theta_i(T_i^{j,l})\sin(\theta_i(T_i^{j,l})). \nonumber
\end{align}
Obviously, each segment requires $2n+1$ sampling points.
The approximate position of the $y$-axis coordinate can be computed in a similar manner.

Considering the error in numerical integration, if we divide the integration interval $[a,b]$ into $n$ intervals, the maximum error of Eq.(\ref{ali:Simpson}) can be expressed as:
\begin{align}
    e_{\max}=\dfrac{(b-a)^5}{180n^4}\max_x|f^{(4)}(x)|. 
\end{align}
It can be found that the integral error is proportional to the fourth derivative of the integrand function. 
Since the integrand function of Eq.(\ref{ali:AppSim}) is higher-order continuous, the error can be kept within an acceptable range. 
We provide further experiments in Sec.\ref{sec:IntegralError} to demonstrate that the integral error does not affect the performance of the planner.

\begin{figure}[b]
    \centering
    \vspace{-0.6cm}
    \setlength{\abovecaptionskip}{-2pt}
    \includegraphics[width=8.5cm]{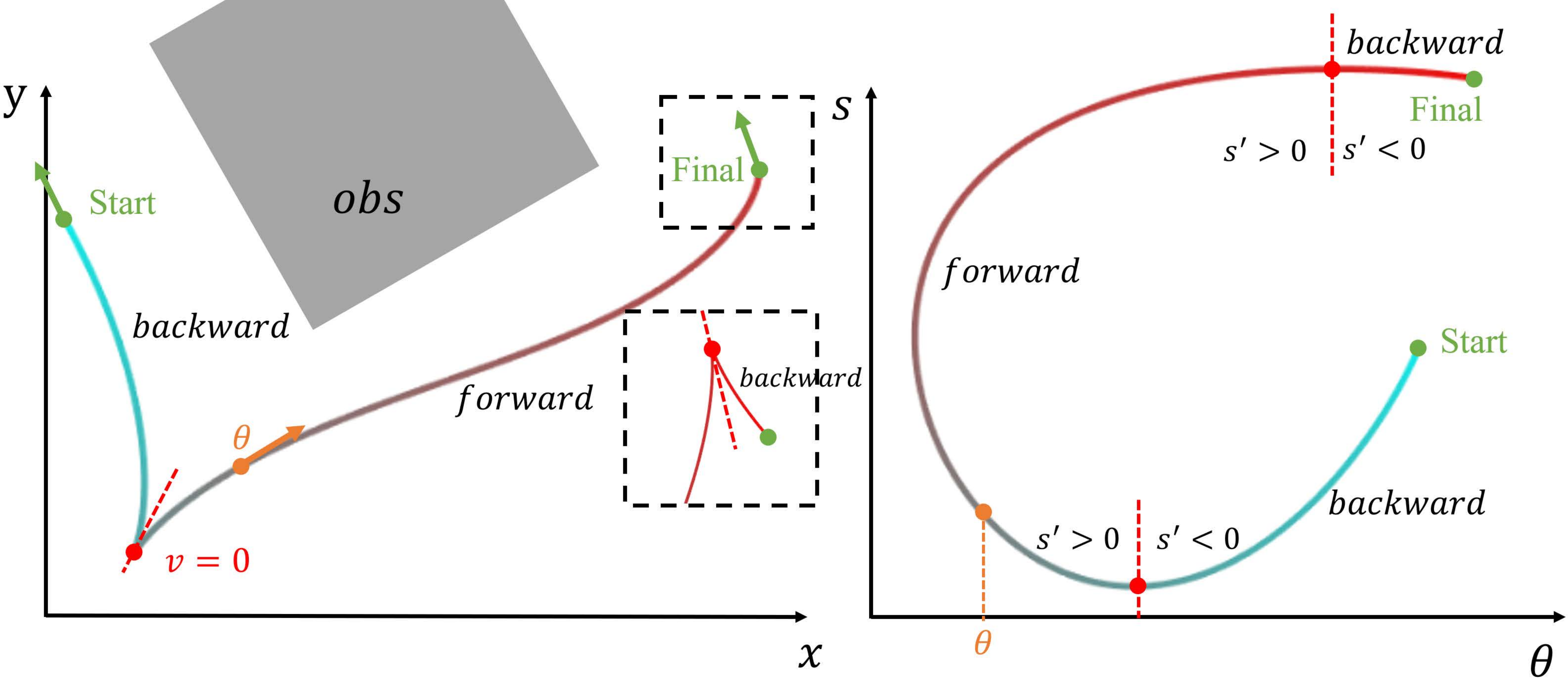}
    \caption{Comparison of the trajectory in $x-y$ space (left) and $\theta-s$ space (right).
             When planning on $x-y$ space with differential flatness, singularities occur when the moving direction changes (red points). 
             However, these can be represented as a smooth trajectory in $\theta-s$ space. }
    \label{fig:comp_diff_cotr}
    % \vspace{-0.7cm}
\end{figure}

Compared to the trajectory based on differential flatness, the MS trajectory has the significant advantage of avoiding the singularity caused by nonholonomic dynamics. 
This means that a unified trajectory can account for both forward and reverse motion, rather than being specified by the planning front-end, as shown in Fig.\ref{fig:comp_diff_cotr}. 
Moreover, lateral slip ($v_y \neq 0$) is also taken into account. 
Compared with directly optimizing control input quantities like OCP, polynomial trajectories can reduce the state propagation error due to their smoother nature and also lower the computational complexity for long trajectories by reducing the optimization variables.

\section{Optimization Problem}\label{section:Optimization_Problem}

In this section, we present the trajectory optimization problem using MS trajectories. 
The constraints for general trajectory optimization and those specific to DD robots will be given respectively in Sec.\ref{sec:constraints}. 
Table.\ref{tab:NotationsOPT} shows the additional notations introduced in this section. 

\begin{table*}[t]
    \caption{Notations of Optimization}
    \vspace{-3pt}
    \centering
    \label{tab:NotationsOPT}
    \fontsize{9}{11}\selectfont % 设置字体大小为12pt，行距为14pt
    % \begin{tabular}{@{}c p{6.5cm}@{}}
    \begin{tabularx}{\textwidth}{@{}c X@{}}  
    \toprule
    ${(\cdot)}^{[i]}, {(\cdot)}^{[i]/[m]}$ & The derivative vectors, where ${(\cdot)}^{[i]}=[ {(\cdot)},{\dot{(\cdot)}},\dots,{(\cdot)}^{(i)}]$ and $ {(\cdot)}^{[i]/[m]}=[ {(\cdot)}^{(m+1)}$,  ${(\cdot)}^{(m+2)},\dots,{(\cdot)}^{(i)}]$. \\ 
    $\boldsymbol\sigma, \boldsymbol \sigma_i(t)$ & The state of the MS trajectory and the polynomial of the $i$th segment, where $\boldsymbol\sigma=[\theta, s]^T$ and $\boldsymbol \sigma_i(t) = [\theta_i(t),s_i(t)]^T$. \\
    $\boldsymbol c=[\boldsymbol c_1^T,... ,\boldsymbol c_M^T]^T$ & The coefficient vector for all $M$ segments, where $\boldsymbol c\in\mathbb R^{2Mh\times2}$ and $h$ is a parameter related to the order of the polynomial. \\
    $\tau_i$ & The corresponding unconstrained optimization variable of $T_i$. \\
    $^w\boldsymbol \sigma'=[\boldsymbol \sigma_1', ... ,\boldsymbol \sigma'_{M-1}]$ & A set representing the final state of each segment, except for the last one, which is called the intermediate points. \\
    $\mathcal C_{d}$, ${\mathcal C_{d}^{i,j}}$, $I_d$ & The constraint, the constraint function for the sampling point of the $j$th sampling interval and the $i$th segment, and the penalty function, where $(\cdot)_d$ represents specific constraints such as kinematic and safety constraints.  \\
    \bottomrule
    \end{tabularx}  
    \vspace{-18pt}
\end{table*}

\subsection{Trajectory Optimization Problem Formulation}\label{subsec:optimization_problem}
Based on the novel representation of trajectories defined above, we construct the optimization problem as:
\begin{subequations}
    \begin{align}
        \mathop{min}\limits_{\boldsymbol c,\boldsymbol T} \; & \mathcal J_0 = \displaystyle\int_0^{T_s}\boldsymbol\sigma^{(h)}(t)^T\boldsymbol W\boldsymbol \sigma^{(h)}(t)dt+\epsilon_TT_s, \label{ali:minJ}\\
        % s.t. & \boldsymbol\psi(t)=\boldsymbol \sigma^{(h)}(t), \forall t\in[0,T_s], \\ 
        s.t. &\boldsymbol \sigma^{[h-1]}(0)=\boldsymbol\sigma_0^{[h-1]}, \label{ali:minJinit}\\
        &\boldsymbol\theta^{[h-1]}(T_s)=\boldsymbol\theta_f^{[h-1]},\boldsymbol s^{[h-1]/[0]}(T_s)=\boldsymbol s_f^{[h-1]/[0]}, \label{ali:minJend}\\
        &\tilde x_f= x_f, \tilde y_f=y_f, \label{ali:minJfinalP}\\
        &\boldsymbol \sigma_{i}^{[h-1]}(T_i)=\boldsymbol \sigma_{i+1}^{[h-1]}(0), \label{ali:minJcont}\\
        &T_s=\sum\limits_{i=1}^{M}T_i,  T_i>0, \label{ali:minJT}\\
        &\mathcal C_{d}(\boldsymbol \sigma(t),\ldots,\boldsymbol \sigma^{(h-1)}(t))\leq0,d\in\mathcal D,t\in[0,T_s]. \label{ali:minJConst}
    \end{align}
\end{subequations}
The quadratic control effort \cite{bertsekas1995dynamic} with time regularization is adopted as a cost functional by Eq.(\ref{ali:minJ}) for motion smoothness \cite{flash1985coordination}, where $\boldsymbol W\in\mathbb R^{2\times 2}$ is a diagonal matrix and $\epsilon_T > 0$ is the weight.
% Since derivatives of $\theta$ and $s$ are common motion states for DD robots, minimizing their higher-order derivatives ensures the smoothness of the motion. 
% $\epsilon_T T_s$, where $\epsilon_T > 0$, is the time regularization term.
Eqs.(\ref{ali:minJinit})-(\ref{ali:minJend}) indicates that the trajectory should satisfy the initial conditions $\boldsymbol\sigma_0^{[h-1]}$ and final conditions $\boldsymbol\theta_f^{[h-1]}, \boldsymbol s_f^{[h-1]/[0]}$. 
Specifically, we set $s_0$ to $0$. 
Eq.(\ref{ali:minJfinalP}) indicates that the trajectory should reach the desired final position $(x_f,y_f)$. 
Eq.(\ref{ali:minJcont}) indicates that the trajectory should be high-order continuous. 
Eq.(\ref{ali:minJT}) defines the total time of the trajectory and constrains the time of each segment to be positive. 
Eq.(\ref{ali:minJConst}) indicates the penalty functions of the additional inequality constraint set $\mathcal D$. 

To solve the optimization problem, we adapt the $\Sigma_{MINCO}$ \cite{wang2022geometrically} trajectory class for our trajectory representation by transforming the optimization variables for this problem.
With the $\Sigma_{MINCO}$ class, the polynomial coefficients $\boldsymbol c$ can be uniquely transformed by the initial state $\boldsymbol \sigma_0^{[h-1]}$, the final state $\boldsymbol \sigma_f^{[h-1]}$, the intermediate points $^w\boldsymbol \sigma'$ and the time $\boldsymbol T$:
% , where $\boldsymbol \sigma_i'$ is the final state of the $i$th segment of the trajectory:
\begin{equation}
    \begin{aligned}[c]
        & \boldsymbol K(\boldsymbol T)\boldsymbol c=[\boldsymbol {\sigma_0^{[h-1]}}, \boldsymbol \sigma_1', \boldsymbol 0,...,\boldsymbol \sigma_{M-1}', \boldsymbol 0, \boldsymbol {\sigma_f^{[h-1]}}]^T, \label{ali:mincoM}
    \end{aligned}
\end{equation}
where $\boldsymbol{K}(T) \in \mathbb{R}^{2Mh \times 2Mh}$ is an invertible banded matrix as given by \cite{wang2022geometrically}, and $\boldsymbol{0} = \boldsymbol{0}_{2 \times (2h-1)}$ is a zero vector.
With Eq.(\ref{ali:mincoM}), we can ensure the polynomial trajectory, represented by $\boldsymbol c, \boldsymbol T$, naturally satisfies the initial and final conditions Eq.(\ref{ali:minJinit})(\ref{ali:minJend}) and continuity Eq.(\ref{ali:minJcont}).

However, the desired destination is generally represented by Cartesian coordinates $x_f,y_f$ instead of the final arc length $s_f$, which means the final state $\boldsymbol {\sigma_f^{[h-1]}}^T$ in Eq.(\ref{ali:mincoM}) is not fixed, specifically the final arc length $s_f$ is not fixed.
The distance that the trajectory travels cannot be determined prior to optimization, and therefore $s_f$ should not be given but rather be part of the optimization variables.
Hence, we need to consider the gradient of the optimization problem with respect to $s_f$. 
We assume that the gradient of the objective function $\mathcal J$ with respect to $\boldsymbol{c}$ has already been computed as $\dfrac{\partial \mathcal J}{\partial \boldsymbol{c}}$. 
As detailed in \cite{wang2022geometrically}, we present the gradient of $\mathcal{J}$ with respect to $s_f$ without proof as:
\begin{align}
    \dfrac{\partial \mathcal J}{\partial s_f}=\boldsymbol{K}^{-T}\dfrac{\partial \mathcal J}{\partial \boldsymbol{c}}\boldsymbol e_{(2M-1)h+1},  
\end{align}
where $\boldsymbol e_i$ is the $i$th column of the unit matrix $\boldsymbol I_{2Mh}\in\mathbb R^{2Mh\times 2Mh}$.  

Here we choose $h=3$ to minimize the linear and angular jerk, thereby achieving smooth trajectories.

\subsection{Inequality Constraints and Gradient Conduction}

For the optimization objective given by Eq.(\ref{ali:minJ}), we can easily compute the gradient. 
Considering the strict positive constraint of time Eq.(\ref{ali:minJT}), we transform the optimization variables from the real time $T_i\in \mathbb{R}^+$ to the unconstrained variables $\tau_i\in \mathbb{R}$ with a smooth bijection: 
\begin{align}
    \tau_i=\left\{\begin{array}{rcl} \sqrt{2T_i-1}-1 & & T_i>1   \\ 1-\sqrt{\frac{2}{T_i}-1} & & T_i\leq1\end{array}. \right. \label{ali:unconsT}
\end{align}

For inequality constraints Eq.(\ref{ali:minJConst}), we can employ penalty methods with a first-order relaxation function $L_1(\cdot)$ \cite{han2023efficient}. 
To approximate the value of the penalty function, we discretely sample the constraint points of the trajectory to constrain it, thus ensuring that the entire trajectory satisfies the constraints:
% \todo{Symbols flooding}
\begin{align}
    % & \mathcal C_d(\boldsymbol c_i,T_i,\hat t,x,y)=\mathcal C_d(\boldsymbol c_i^T\boldsymbol\beta(\hat t T_i),...,\boldsymbol c_i^T\boldsymbol\beta^{(h)}(\hat t T_i),x,y), \nonumber\\
    % & \mathcal C_d(\boldsymbol c,T,\hat t,x,y)=\mathcal C_d(\boldsymbol c^T\boldsymbol\beta(\hat t T),...,\boldsymbol c^T\boldsymbol\beta^{(h)}(\hat t T),x,y), \nonumber\\
    & \mathcal C_d^{i,j}(\boldsymbol c,\boldsymbol T)=\mathcal C_d(\boldsymbol c_i^T\boldsymbol\beta(\frac{j}{n}T_i),...,\boldsymbol c_i^T\boldsymbol\beta^{(h-1)}(\frac{j}{n}T_i),x^{i,j},y^{i,j}), \nonumber\\
% \end{align}
% \begin{align}
    & I_d(\boldsymbol c,\boldsymbol T) = \varsigma_d\sum\limits_{i=1}^M\sum\limits_{j=0}^{n}\dfrac{T_i}{n}\overline\nu_jL_1(\mathcal C_d^{i,j}(\boldsymbol c,\boldsymbol T)),
    \label{ali:sample}
\end{align}
where $\varsigma_d$ is the weight of the constraint $d$. 
$(\nu_0, \nu_1, \ldots , \nu_{n-1}, \nu_{n}) = (0.5, 1, \ldots , 1, 0.5)$ are the coefficients from the trapezoidal rule. 
Here the number of samples $n$ is the same as the integration intervals in Eq.(\ref{ali:AppSim}), which avoids over-calculation. 
However, the coefficients $\boldsymbol c_i$ and the time $T_i$ can only represent the current motion states and their higher-order derivatives, but not the current position. 
That is because the position is the result of the integration, which depends on the previous motion states. 
Therefore, the constraint function $\mathcal C_d^{i,j}$ for the sampling point is also defined as a function of $x^{i,j},y^{i,j}$. 

Finally, the optimization problem can be reformulated from Eq.(\ref{ali:mincoM})(\ref{ali:unconsT})(\ref{ali:sample}) as:
\begin{equation}
    \begin{aligned}[c]
        \min\limits_{^w\boldsymbol \sigma',\boldsymbol \tau,s_f} \mathcal J_s = & \mathcal J_0(\boldsymbol c(^w\boldsymbol \sigma',\boldsymbol \tau, s_f),\boldsymbol T(\boldsymbol \tau)) \\
        &\;\;\;\;\;\;\;\;\; + \sum\boldsymbol I_d(\boldsymbol c(^w\boldsymbol \sigma',\boldsymbol \tau, s_f),\boldsymbol T(\boldsymbol \tau)).
        \label{ali:noncons}
    \end{aligned}
\end{equation}

Without loss of generality, we give the gradient of the constraint function $\mathcal C_d^{i,j}$ with respect to the optimized variables based on the chain rule, and for simplicity, we omit subscripts:
\begin{align}
    \dfrac{\partial\mathcal C_d}{\partial\boldsymbol  c}=\sum\limits^{h-1}_{\overline h=0}\boldsymbol\beta^{(\overline h)}(\dfrac{\partial \mathcal C_d}{\partial\boldsymbol \sigma^{(\overline h)}})^T+\dfrac{\partial x}{\partial\boldsymbol c}\dfrac{\partial \mathcal C_d}{\partial x}+\dfrac{\partial y}{\partial\boldsymbol c}\dfrac{\partial \mathcal C_d}{\partial y}, 
\end{align}
\begin{align}
    \dfrac{\partial \mathcal C_d}{\partial\boldsymbol T}=\sum\limits^{h-1}_{\overline h=0}\dfrac{\partial\boldsymbol\sigma^{(\overline h)}}{\partial\boldsymbol T}\dfrac{\partial \mathcal C_d}{\partial\boldsymbol\sigma^{(\overline h)}}+\dfrac{\partial x}{\partial\boldsymbol T}\dfrac{\partial \mathcal C_d}{\partial x}+\dfrac{\partial y}{\partial\boldsymbol T}\dfrac{\partial \mathcal C_d}{\partial y}.  
\end{align}

Thus, we can give the derivative of the constraint function $I_d$:
\begin{align}
    & \dfrac{\partial I_d}{\partial \boldsymbol  c} = \varsigma_d\sum\limits_{i=1}^M\sum\limits_{j=0}^{n}\dfrac{T_i}{n}\overline\nu_j\dfrac{\partial \mathcal C_d}{\partial \boldsymbol c}\dfrac{\partial L_1(\mathcal C_d)}{\partial \mathcal C_d}, \\
    & \dfrac{\partial I_d}{\partial\boldsymbol T} = \varsigma_d\sum\limits_{i=1}^M\sum\limits_{j=0}^{n}\overline\nu_j(
    \dfrac{1}{n}L_1(\mathcal C_d)e_i+
    \dfrac{T_i}{n}\dfrac{\partial \mathcal C_d}{\partial\boldsymbol T}\dfrac{\partial L_1(\mathcal C_d)}{\partial \mathcal C_d}),  
\end{align}
where $e_i$ is the $i$th column of the identity matrix $I_M\in\mathbb R^{M\times M}$. 
The gradients of $\boldsymbol \sigma^{(\overline h)}$ with respect to $\boldsymbol c$ and $\boldsymbol T$ can be easily computed. 
However, $x,y$ involve integrals of $\boldsymbol \sigma$ and $\boldsymbol{\dot\sigma}$, so they have gradients for all previous sampling points. 
For example, the gradient of the penalty function $\mathcal C_d^{i,j}$ with respect to $x^{i,j}$ is $g_x^{i,j}$ at a sampling point, which corresponds to the point after the $j$th sampling interval of $i$th segment, and we can get its gradient with respect to $\boldsymbol c$ from Eq.(\ref{ali:AppSim}) as:
\begin{align}
    & \Gamma_{i,j} = \hat x_{i}^{j,0}+4\hat x_{i}^{j,1}+\hat x_{i}^{j,2}, \nonumber\\
    & \dfrac{\partial x^{i,j}}{\partial \boldsymbol c_m}\dfrac{\partial\mathcal C_d}{\partial x^{i,j}} = g_x^{i,j}(\dfrac{T_m}{6n}\sum\limits_{k=1}^{n}\dfrac{\partial\Gamma_{m,k}}{\partial \boldsymbol c_m}), m\in[1,i-1], \nonumber\\
    & \dfrac{\partial x^{i,j}}{\partial \boldsymbol c_i}\dfrac{\partial\mathcal C_d}{\partial x^{i,j}} = g_x^{i,j}(\dfrac{T_i}{6n}\sum\limits_{k=1}^{j}\dfrac{\partial\Gamma_{i,k}}{\partial \boldsymbol c_i}). \label{ali:x2c}
\end{align}

Similarly, we can get the gradient with respect to $\boldsymbol T$ as:
\begin{align}
    & \dfrac{\partial x^{i,j}}{\partial T_m}\dfrac{\partial\mathcal C_d}{\partial x^{i,j}} = 
    g_x^{i,j}\sum\limits_{k=1}^{n}(\dfrac{\Gamma_{m,k}}{6n}  +\dfrac{T_m}{6n}\dfrac{\partial\Gamma_{m,k}}{\partial T_m}),m\in[1,i-1]
    , \nonumber\\
    & \dfrac{\partial x^{i,j}}{\partial T_i}\dfrac{\partial\mathcal C_d}{\partial x^{i,j}} = 
    g_x^{i,j}\sum\limits_{k=1}^{j}(\dfrac{\Gamma_{i,k}}{6n}  + \dfrac{T_i}{6n}\dfrac{\partial\Gamma_{i,k}}{\partial T_i}). \label{ali:x2T}
\end{align}

For convenience, we compute the gradient of the sampled point $\dfrac{\partial\Gamma_{i,j}}{\partial T_i}$ and $\dfrac{\partial\Gamma_{i,j}}{\partial \boldsymbol c_i}$ from Eq.(\ref{ali:AppSim}) and save them in vectors while integrating to compute the whole trajectories. 
Therefore, we only need to compute the gradient of the constraint function on $\boldsymbol \sigma^{(\overline h)}, x$ and $y$, after which it can be transferred to the optimized variables using the chain rule.

\subsection{Final Position Constraints}\label{subsec:finalPos}
Unlike trajectories based on differential flatness in the Cartesian frame, the final position of the MS trajectory is computed by integral approximation and could result in a significant difference from the desired destination.
Therefore, it is necessary to add final position constraints specifically to reduce the error between the final position of the trajectory $(\tilde{x}_f,\tilde{y}_f)$ and the desired position $(x_f,y_f)$. 
Using the integral expression Eq.(\ref{ali:AppSim}), we can easily formulate the constraint for the x-axis as:
\begin{align}
    \mathcal C_{fx}(\boldsymbol{c},\boldsymbol{T})=\tilde{x}_f(\boldsymbol{c},\boldsymbol{T}) - x_{f}. 
\end{align}

Unlike other constraints, final position constraints are generally given with a maximum error $e_{max}$. 
Although we can assign a large penalty weight to guarantee the accuracy, it may change the topology of the trajectory or slow down the convergence. 
Moreover, if we only use the penalty function method, the optimization results may not meet the required accuracy.
To iteratively reduce this error, we introduce the Powell-Hestenes-Rockafellar Augmented Lagrangian Methods (PHR-ALM) \cite{hestenes1969multiplier,powell1969method,rockafellar1974augmented}. 
We define $\mathcal J'$ as an unconstrained optimization problem in Eq.(\ref{ali:noncons}) excluding the final position constraints and transformed with the penalty function. 
The new optimization problem can be expressed as:
\begin{equation}
    \begin{aligned}[c]
        \mathcal J_\rho(^w\boldsymbol \sigma', & \boldsymbol \tau,s_f,\boldsymbol \lambda) := \mathcal J_s'\\
        & +\sum\limits_{\iota=x,y}\dfrac{\rho}{2}\|\mathcal C_{f\iota}(\boldsymbol c(^w\boldsymbol \sigma',\boldsymbol \tau, s_f),\boldsymbol T(\boldsymbol \tau))+\dfrac{\lambda_\iota}{\rho}\|^2, \label{align:ALMJ}
    \end{aligned}
\end{equation}
where $\boldsymbol\lambda$ is the dual variable and $\rho>0$ is the weight of the augmentation term. 
Next, it can be updated as follows:
\begin{subequations}\label{ali:iter}
    \begin{numcases}{}
        {^w\boldsymbol \sigma'}^{k+1},\boldsymbol \tau^{k+1},s_f^{k+1} = \operatorname{argmin}\mathcal J_\rho(^w\boldsymbol \sigma',\boldsymbol \tau,s_f,\boldsymbol \lambda^k), \notag\\
        \lambda^{k+1}_\iota=\lambda_\iota^k+\rho^k \mathcal C_{f\iota} , \;\;\iota=x,y,          \qquad\qquad\qquad\qquad\;\;\;\;\;\;\text{(\ref{ali:iter})}  \notag\\
        \rho^{k+1} = \min[(1+\varrho)\rho^k, \rho_{max}], \notag
    \end{numcases}
  \end{subequations}
where $\varrho>0$ ensures that $\rho^k$ is an increasing non-negative sequence, and $\rho_{max}$ is used to avoid $\rho^k$ becoming too large. 
L-BFGS\cite{LBFGS} is used to solve the optimization problem $\operatorname{argmin}\mathcal J_\rho$. 
The iterative calculation according to Eq.(\ref{ali:iter}) is performed until the following condition is met:
\begin{align}
    \sqrt{(\tilde{x}_f-x_f)^2+(\tilde{y}_f-y_f)^2}<e_{max}, \label{align:finalPos_emax} 
\end{align}
i.e., we can gradually control the error of the final position within the set range $e_{max}$.

\section{Constraints of DD Robots} \label{sec:constraints}
In this section, we focus on the specific form of $\mathcal C_d$ in Eq.(\ref{ali:minJConst}), which fundamentally ensures the efficient and safe operation of differential drive robots. 
To ensure the dynamic feasibility of the trajectory, we introduce velocity and acceleration constraints in Sections \ref{subsec:vel} and \ref{subsec:acc}.
The safety constraint, discussed in Section \ref{subsec:safety}, is used to prevent collisions. 
Additionally, the Segment Duration Balance constraint, introduced in Section \ref{subsec:uniform_time}, serves as an auxiliary constraint to improve the numerical properties of the optimization process.

\subsection{Coupled Linear Velocity and Angular Velocity}\label{subsec:vel}
%Considering the kinematic constraints of the robot, it is necessary to keep the velocity and angular velocity within reasonable limits. 
Different from Ackerman or quadrotor platforms, the linear velocity and angular velocity of DD robots are closely coupled as both are produced from the two driving wheels.
Any changes in the left or right wheels can affect the forward speed or turning rate simultaneously.
Therefore, it is necessary to consider the constraints of linear and angular velocities with the relation constraint, instead of linear and angular velocities separately. 
Considering that calculating an accurate $\omega_{\max}-v_x$ profile by dynamics is inefficient, we approximate it with the kinematic model. 
% Although considering the dynamic coupling relationship between ground contact is complex, we can approximate it using a kinematic model. \todo{elaborate the 'complex', ground contact, right or not?}
% \todo{repetitive or not logical, rewrite. }
Assuming that the maximum velocity of the two wheels is fixed, the maximum angular velocity versus velocity can be obtained using Eq.(\ref{ali:omegaz})(\ref{ali:vx}):
\begin{align}
    \omega_{max}(v_x)=\dfrac{2(v_{M}-|v_x|)}{y_{Il}-y_{Ir}},
    \label{ali:maxvomega}
\end{align}
where $\omega_{max}$ is the current maximum angular velocity, and $v_M$ is the maximum velocity when moving forward in a straight line. 
Additionally, we assume $x_{Iv}=0$. 
However, a more common scenario is that the maximum angular velocity of the robot cannot reach the maximum value calculated by Eq.(\ref{ali:maxvomega}). 
% \todo{In practice, the angular velocity of rotating in place ($v=0$) often does not reach the theoretical maximum.}
% , and the rotational speeds of the two wheels at this time may be much less than maximum linear velocity. 
% Therefore, we simplify based on this assumption. \todo{what assumption}
We still assume that the maximum value of angular velocity and linear velocity are linearly related. 
If the maximum angular velocity when $v_x=0$ is $\omega_M$, we can rewrite Eq.(\ref{ali:maxvomega}) as:
\begin{align}
    \omega_{max}(v_x)=\dfrac{v_{M}-|v_x|}{v_M}\omega_M.
\end{align}

Considering that some robots have extra limitations when reversing, i.e., the maximum reverse velocity may be less than the maximum forward velocity, we constrain them separately. 
Assuming that the maximum reverse velocity is $v_{\overline M}\leq0$, we can give the velocity and angular velocity constraints: 
\begin{align}
    &\mathcal C_{m+}(\boldsymbol{\dot\sigma})=\eta_\omega\omega v_M+\omega_Mv_x-v_M\omega_M, \eta_w\in\{-1,1\},\;\; \label{align:moment} \\ 
    &\mathcal C_{m-}(\boldsymbol{\dot\sigma})=-\eta_\omega\omega v_{\overline M}-\omega_Mv_x+v_{\overline M}\omega_M, \eta_w\in\{-1,1\}, \nonumber      
\end{align}
where $\mathcal C_{m+}$ and $\mathcal C_{m-}$ are constraints for forward and reverse, respectively. 
Specifically, we can set $v_{\overline M}=0$ to prohibit reversing. 

Although we make an assumption of a linear relationship, more precise constraints can be determined in real experiments using any differentiable constraint functions if necessary.

\subsection{Linear Acceleration and Angular Acceleration}\label{subsec:acc}
To smooth the trajectory of the robot and to take into account the moment limitations of the actuator, we also set constraints on the linear acceleration $a$ and angular acceleration $\alpha$ of the robot.
However, calculating time-varying acceleration limits through dynamics is unnecessary. 
We simplify by setting a constant maximum value for each of them, and their corresponding constraint functions are: 
% Or We set a constant maximum value for each of them, and their corresponding constraint functions are: 
% \todo{ give the reason for 'complex'. not reasonable just from a word of 'complex'. same for the 'complex' above}
\begin{equation}
    \begin{aligned}[c]
        & \mathcal C_a(\boldsymbol{\ddot\sigma}) = a^2-a_M^2, \\ 
        & \mathcal C_\alpha(\boldsymbol{\ddot\sigma}) = \alpha^2-\alpha_M^2, 
        \label{align:acceleration}
    \end{aligned}
\end{equation}
where $a_M$ and $\alpha_M$ are the maximum values of linear and angular acceleration, respectively.

\subsection{Safety}\label{subsec:safety}
We use the Euclidean Signed Distance Field (ESDF) to constrain the distance from the robot to obstacles. 
% After getting the gridmap, we can compute the ESDF using an efficient $O(r)$ algorithm \cite{felzenszwalb2012distance}, where $r$ is the number of updated grids.
% Because of the round-off errors of the ESDF introduced by the discretization of the gridmap, bilinear interpolation is used to improve the accuracy of the distance and gradient information.
We take $\gamma_b$ points $\chi_{b,l} \in \mathbb{R}^2$ along the profile of the robot. 
$\chi_{b,l}$ are the coordinates of the profile points in the body frame and are specified before the optimization.
Then their positions in the world frame can be calculated:
\begin{align}
    \boldsymbol p_{b,l}(x,y,\theta)=[x,y]^T+\boldsymbol R(\theta) \boldsymbol\chi_{b,l}, l\in\{1,2,...,\gamma_b\},
\end{align}
where $x,y$ are the position of the origin of the body frame and $\boldsymbol R(\theta)$ is the rotation matrix of the robot. 
By ensuring the ESDF values of the contour points are greater than the safe distance $d_s$, we can prevent the robot from colliding with obstacles. 
Thus, the safety constraints can be expressed as:
\begin{align}
    \mathcal C_{s,l}(x,y,\theta)=d_s-E(\boldsymbol p_{b,l}(x,y,\theta)), l\in\{1,2,...,\gamma_b\},
    \label{ali:safe}
\end{align}
where $E(\cdot)$ is the ESDF value got by bilinear interpolation. 
In complex environments or where the robot is asymmetric, we can choose to constrain more contour points, thus ensuring full utilization of the safe space. 
Conversely, we can choose to constrain only one point $\chi_b=[0,0]^T$, i.e., model the robot as a circle, thus improving the efficiency of the optimization.

\subsection{Segment Duration Balance} \label{subsec:uniform_time}
During the optimization process, the time duration of the segments could become unbalanced.
If the time $T_i$ of segment $i$ is too long, it will result in larger intervals between the sampling points, which can hinder the accurate approximation of the sampling constraint function and increase the error in the integration. 
Therefore, we aim to keep the time allocated to each segment within a certain range of deviation, ensuring that they do not vary excessively from one another. 
For the $i$th segment, we give the balanced duration constraint:
\begin{equation}
    \begin{aligned}[c]
        & \mathcal C_{Tlow}(T_i)=\epsilon_{low}\frac{1}{M}\sum_{m=1}^M T_m - T_i, \\ 
        & \mathcal C_{Tupp}(T_i)=T_i - \epsilon_{upp}\frac{1}{M}\sum_{m=1}^M T_m, 
        \label{ali:unitime}
    \end{aligned}
\end{equation}
% \todo{More detailed variable explanations are needed. }
where $\epsilon_{low}\in[0,1)$ is the lower limit and $\epsilon_{upp}>1$ is the upper limit. 
With Eq.(\ref{ali:unitime}), each segment's time is constrained within a certain range of the average, which ensures that the duration of each segment is neither too long nor too short.

\section{Planning and control system design} \label{sec:navi_sys}

\begin{figure}[b]
    \centering
    \vspace{-0.5cm}
    \setlength{\abovecaptionskip}{3pt}
    \includegraphics[width=8.8cm]{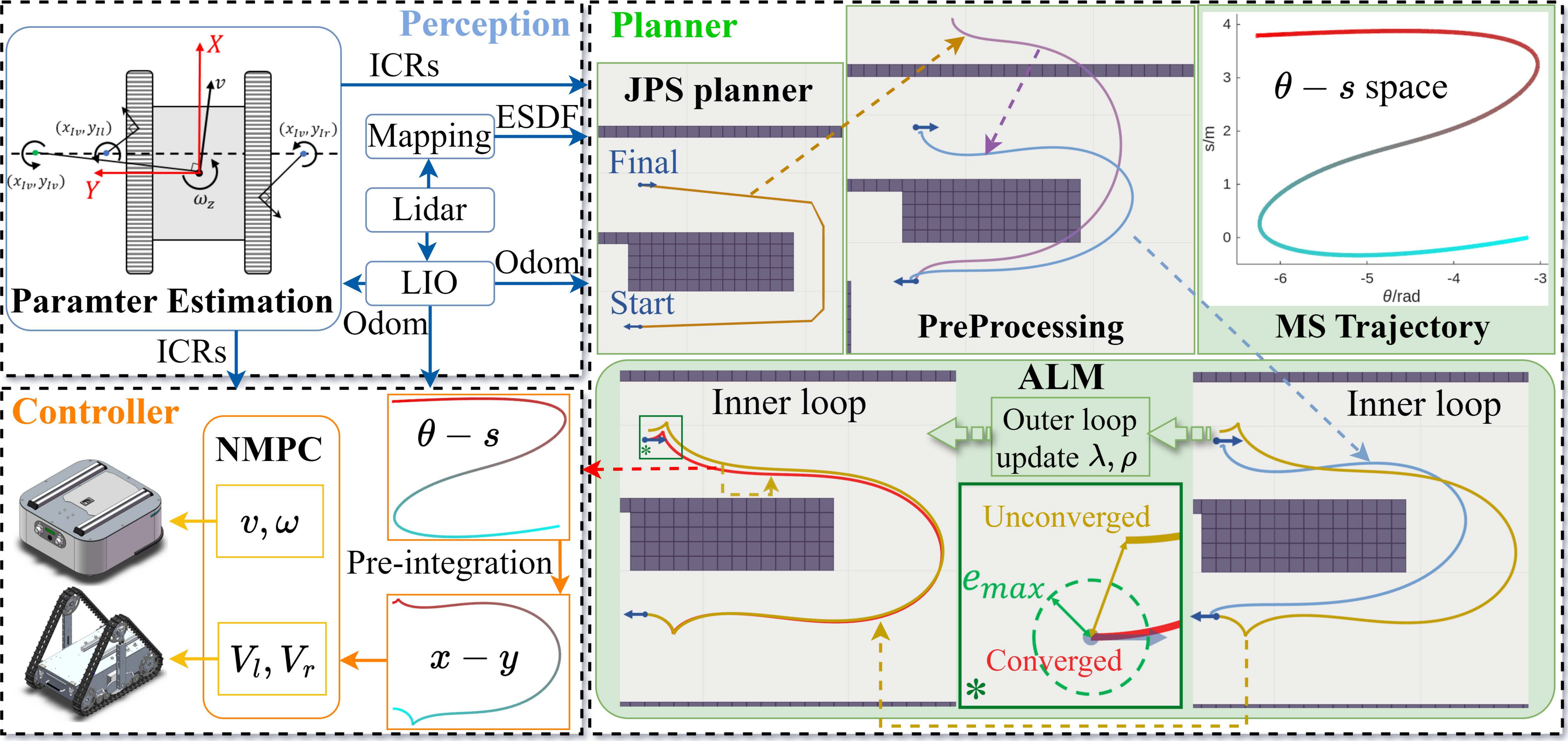}
    % Overview of the proposed planning and control system. 
    % Perception: 系统使用基于Lidar-Inertial Odometry的SLAM，并获得ESDF以及进行参数估计
    % Planner: 系统使用JPS获得全局路径，并在路径上采样MS轨迹的intermediate points以获得初始轨迹。轨迹前处理被用于获得与路径更相近的轨迹。ALM方法通过迭代优化以获得满足运动学和安全约束的轨迹，并保证机器人到达期望的终点。

    % Controller: 系统对MS轨迹进行预积分以获得参考轨迹，并根据机器人的驱动方式选择对应的NMPC进行控制。
    % Perception: The system employs SLAM based on Lidar-Inertial Odometry, which enables ESDF and parameter estimation.
    % Planner: The system uses JPS to get the global path and samples intermediate points to generate the initial MS trajectory. Trajectory pre-processing is applied to ensure close alignment with the global path. The ALM method iteratively optimizes the trajectory to satisfy constraints and reach the desired final positions.
    % Controller: The system uses pre-integration on the MS trajectory to derive the reference trajectory. Depending on the driving method, the appropriate NMPC is selected for control.
    \caption{
        Overview of the a example navigation system with our planner. 
        In perception, lidar is used for Lidar-Inertial Odometry (LIO) and mapping.
        With the result of odometry and ESDF map, our planner uses Jump Point Search (JPS) to get the global path (brown) and to generate the initial MS trajectory (purple). 
        Trajectory pre-processing (blue) ensures close alignment of the trajectory with the global path. 
        The Augmented Lagrangian Method (ALM) iteratively optimizes the trajectory to satisfy constraints (yellow) and reach the desired final positions (red).
        The system uses pre-integration to get the reference trajectory and selects appropriate NMPC controller depending on the driving method.
    }
    \label{fig:Block_Diagram}
    % \vspace{-0.5cm}
\end{figure}

Building upon the trajectory optimization mentioned above, we further complete a full planning and control system for the differential drive robot class. 
We introduce a global path search to provide initial values and implement trajectory preprocessing to reduce topological changes and enhance robustness. 
We also introduce parameter estimation to estimate the kinematic parameters of the robot and develop a controller based on these parameters. 
The system diagram is shown in Fig.\ref{fig:Block_Diagram}. 

\subsection{Global Path Search}
We use Jump Point Search (JPS) \cite{harabor2011online} as our global planner to find a path from the initial position to the final position.
JPS can generate global paths quickly, which is important for fast replanning scenarios. 

\subsection{Trajectory Preprocessing}
\begin{figure}[t]
    \centering
    % \vspace{-0.5cm}
    % \setlength{\abovecaptionskip}{-0.1cm}
    \includegraphics[width=8.8cm]{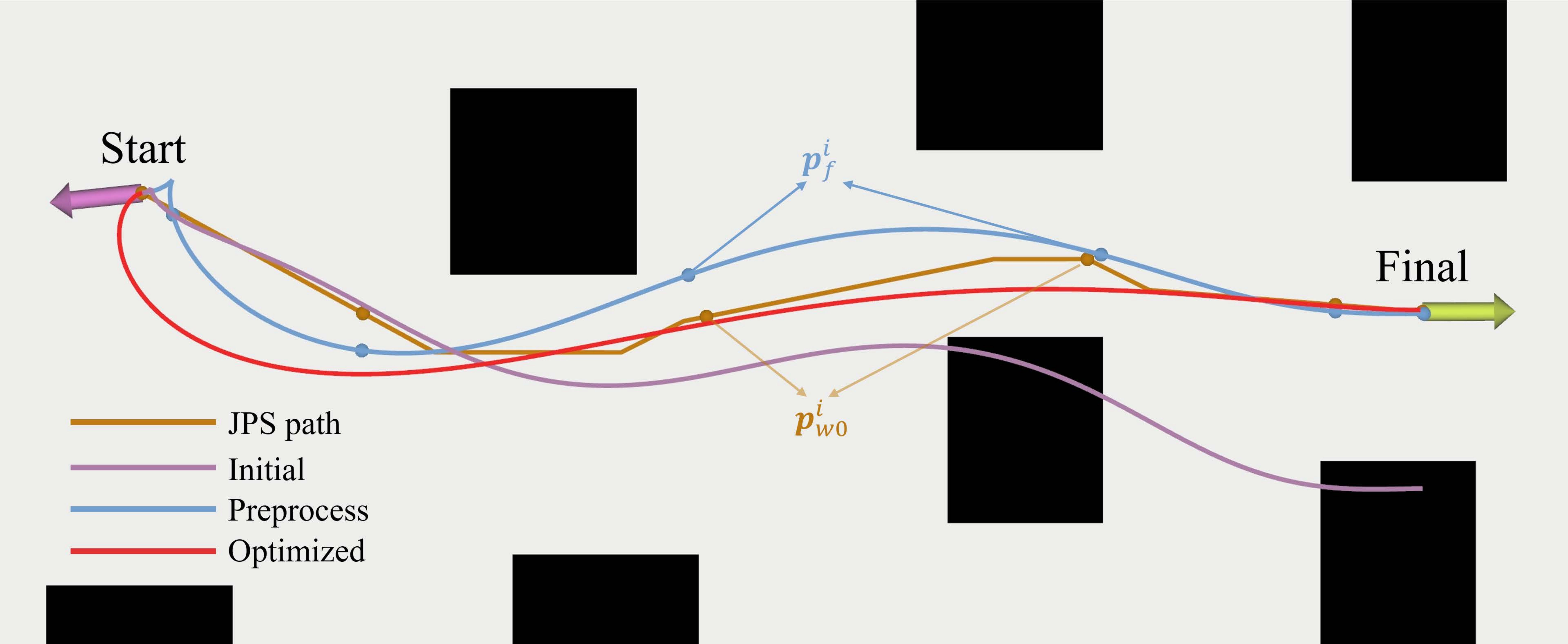}
    \caption{Planning result. 
    % 由于获得优化初值时的误差，初始轨迹可能在障碍物里面（紫）。利用轨迹预处理（蓝）我们可以在优化前使轨迹尽可能贴近前端路径（蓝），从而避免优化失败。优化后的轨迹（红）可以更好的满足约束条件。
    Due to errors in getting the initial values, the initial trajectory may be inside obstacles (purple). 
    By using trajectory preprocessing (blue), we can make the trajectory as close as possible to the front path (brown) before optimization, thus avoiding optimization failure. 
    The optimized trajectory (red) can better satisfy the constraints.
    }
    \label{fig:pathSDF}
    \vspace{-0.5cm}
\end{figure}

Considering that JPS does not consider any kinematic information, sampling on the global path to directly get the initial values of the trajectory is not reasonable. 
The global path is composed of coordinate points $\{\boldsymbol{p}_g\}\in\mathbb R^2$ without orientation, which will result in poor initial orientation and may lead to a significant difference between the initial trajectory and the global path, as shown in Fig.\ref{fig:pathSDF}.
The lack of kinematic information may also lead to poorly assigned velocities. 
Therefore, it is necessary to perform preprocessing to get better initial values before trajectory optimization Eq.(\ref{align:ALMJ}). 
After getting the global path and sampling to get the initial values, we can achieve better initial values through preprocessing with a simpler optimization.  

When using JPS to get the initial trajectory, we can also determine the initial value of the final position for each trajectory segment, expressed as $\boldsymbol p_{w0}\in\mathbb R^{M\times 2}$.
We apply constraints on the final position of each segment: 
\begin{align}
    \mathcal I_{sf}(\boldsymbol p^i_{f}(\boldsymbol c, \boldsymbol T))=\sum_{i=1}^M(\boldsymbol p_f^i-\boldsymbol p_{w0}^i)^2, 
    \label{align:Isf}
\end{align}
where $\boldsymbol p_f^i$ is the integrated final position of the $i$th segment, and $\boldsymbol p_{w0}^i$ is the initial final position of the $i$th segment. 
Equation (\ref{align:Isf}) ensures that the trajectory's topology and success rate are not adversely affected by poor initial values, by bringing the processed trajectory closer to the global path. 
Additionally, it ensures the trajectory's integral final position is closer to the desired final position. 

Similar to Sec.\ref{subsec:optimization_problem}, we design the optimization problem as:
\begin{equation}
    \begin{aligned}[c]
        \min\limits_{^w\boldsymbol \sigma_p',\tau_p,s_f} \; J_{pre}  = & J_0(\boldsymbol c(^w\boldsymbol \sigma_p',\boldsymbol \tau_p),\boldsymbol T_p(\boldsymbol \tau_p)) \\
        & \;\;\;\;\;\;+ \sum\boldsymbol I_{d'}(\boldsymbol c(^w\boldsymbol \sigma_p',\boldsymbol \tau_p),\boldsymbol T_p(\boldsymbol \tau_p)),\label{ali:pre_process}
    \end{aligned}
\end{equation}
where $d'=\{m,a,\alpha,Tlow,Tupp,sf\}$. 
By considering Eq.(\ref{align:moment}), Eq.(\ref{align:acceleration}), Eq.(\ref{ali:unitime}), and Eq.(\ref{align:Isf}) via the penalty function method, we can generate a trajectory that is closer to the initial global path and more dynamically feasible. 
Here we set a large convergence tolerance to reduce its computation time. 

\subsection{Replanning Strategy} \label{subsec:replanning_strategy}
\textbf{Starting Positions of Replanning: }
Considering the computation time of the planner, we set a time interval $T_R$ as the expected computation time. 
At time $t_c$, when the robot initiates replanning, we select the position $\boldsymbol{p}_R$ at time $t_c+T_R$ as the starting position for replanning. 
As long as the planner completes the computation within $T_R$ and transmits the trajectory to the controller, the controller can replace the old trajectory with the new one at $t_c+T_R$, ensuring the smoothness of the trajectory.
% Here we set $T_R=20\text{ms}$, and in almost all cases, the planner can complete the computation. 
% The time consumed by planner is much less than the computation time of mapping and actuator delay, thereby minimizing the impact of the planner on the system's response speed and enhancing the robot's applicability in time-sensitive scenarios.

\textbf{Selection of Starting Positions of JPS: } 
Considering that changes in topology can greatly impact the current motion of the robot, we do not use the robot's current position as the starting position for replanning.
Instead, we search forward along the current trajectory for a certain period of time $T_J$ and get the last safe point $\boldsymbol{p}_J$ as the starting position of JPS, thus reducing topology changes. 
It is worth noting that $\boldsymbol{p}_J$ is different from $\boldsymbol{p}_R$. 
The path from $\boldsymbol{p}_R$ to $\boldsymbol{p}_J$ will be connected to the JPS path as part of the global path.

\textbf{Truncation: }
Since trajectory optimization requires more computation time compared to the front-end, we truncate the reference paths whose lengths are over $l_{max}$ and use the front part for optimization. 
Generally, replanning does not need to precisely reach the interim destination. 
We relax the final position constraint during replanning, i.e., we set a large convergence condition $e_{max}$ in Eq.(\ref{align:finalPos_emax}), thus saving the replanning time. 

\subsection{Parameter Estimation}\label{subsec:Parameter_estimation}
In Sec.\ref{sec:Kinematic_Model}, we introduce ICRs to describe the kinematic model for the different DD robots.
To improve accuracy, it is necessary to perform online estimation for ICRs.
Inspired by \cite{pentzer2014model}, we use the Extended Kalman Filter (EKF) to estimate the position of the ICRs. 
The ICRs are modeled as constants disturbed by random noise. 
We denote the state of the robot as $\boldsymbol {ST}=[x,y,\theta,y_{Il},y_{Ir},x_{Iv}]^T$ and express the kinematic as follows:
\begin{align}
    \begin{bmatrix}
        \dot x \\  \dot y \\   \dot \theta \\   \dot y_{Il} \\   \dot y_{Ir} \\     \dot x_{Iv} \\
        \end{bmatrix} = 
        \begin{bmatrix}
        v_x\cos\theta-v_y\sin\theta+\Upsilon_x \\
        v_x\sin\theta+v_y\cos\theta+\Upsilon_y \\
        \omega + \Upsilon_\omega \\
        \Upsilon_{y_{Il}}\\
        \Upsilon_{y_{Ir}}\\
        \Upsilon_{x_{Iv}}\\
    \end{bmatrix}, 
    \label{ali:track_EKF}
\end{align}
where $\Upsilon_\cdot$ denotes zero-mean Gaussian noise. 
We use Eq.(\ref{ali:track_EKF}) as the state propagation model. 
Given that the robot's position can be directly obtained from the localization module, we set the observation equation to $\boldsymbol {OB}=[x_{ob},y_{ob},\theta_{ob}]^T$. 
Based on the observation equation, we can calculate the filter gain and update the state and covariance.

Considering that we also need EKF for state estimation, we can perform state estimation simultaneously to obtain the robot's current kinematic parameters. 

\subsection{Controller design}
Given the robot's kinematics Eq.(\ref{ali:omegaz}-\ref{ali:vy}), and the reference trajectory given in Sec.\ref{section:Optimization_Problem}, we use Nonlinear Model Predictive Control (NMPC) by solving the optimal control problem in a receding horizon manner. 
Generally, the localization module provides only position information, so we use the position as the state of NMPC rather than directly using motion states.
NMPC minimizes the error between the predicted state and the reference state for a finite forward horizon $[t_0,t_0+T_h]$, where $t_0$ is the current time and $T_h$ is the prediction horizon. 
We set a fixed time step $dt$ and the corresponding horizon length $N=T_h/dt$ to discretize the trajectory. 
Here, the timestep $i$ represents time $t_i=t_0+i\,dt$. 
Subsequently, the optimal control problem is transformed into the following nonlinear programming:
\begin{align}
    \footnotesize
    \boldsymbol {U^*} & =\arg\min_{\boldsymbol{u}}\sum_{i=k}^{k+N-1}(\overline{\boldsymbol p}_i^T\boldsymbol {W}_p \overline{\boldsymbol p}_i+\overline{\boldsymbol  u}_i^T\boldsymbol {W}_u \overline{\boldsymbol  u}_i)+\overline{\boldsymbol p}_{N_k}^T\boldsymbol {W}_p \overline{\boldsymbol p}_{N_k}, \nonumber\\
    & s.t. \;\;\; \boldsymbol p_{i+1}=f(\boldsymbol p_i,\boldsymbol  u_i),\boldsymbol  u_{\min}<\boldsymbol  u_{i}<\boldsymbol  u_{\max}, 
\end{align}
where $k$ is the current timestep, $\boldsymbol p_i, \boldsymbol u_i$ are the state and input of timestep $i$, 
$\overline{\boldsymbol p}_i=\boldsymbol p^r_i-\boldsymbol p_i, \overline{\boldsymbol  u}_i=\boldsymbol  u^r_i-\boldsymbol  u_i$ are the state and input errors, $\overline{\boldsymbol p}_{N_k} \equiv \boldsymbol p^r_{k+N}-\boldsymbol p_{k+N}$ is the final-state error, 
${\boldsymbol p}^r_i, {\boldsymbol u}^r_i$ are the reference state and input of timestep $i$ calculated from the optimized trajectory. 
$\boldsymbol  W_p$ and $\boldsymbol  W_u$ are the weight matrices of the corresponding terms. 
We solve this NMPC problem using ACADO \cite{houska2011acado} with qpOASES \cite{ferreau2014qpoases}. 
For the robot controlled via two-wheel speed, we set the differential equation as:
\begin{align}
    \dot{\boldsymbol p}=\begin{bmatrix}
        v_x\cos\theta-v_y\sin\theta \\
        v_x\sin\theta+v_y\cos\theta \\
        \omega \\
        \end{bmatrix},
\end{align}
where $v_x,v_y,\omega$ are the robot's kinematics corresponding to Eq.(\ref{ali:omegaz}-\ref{ali:vy}), and the robot's kinematic parameters are estimated in real time by Sec.\ref{subsec:Parameter_estimation}. 

For the robot controlled by linear velocity $v_x$ and angular velocity $\omega$, we set the differential equation as: 
% \todo{why don't keep using s $\theta$ v w ?}
\begin{align}
    \dot{\boldsymbol p}=\begin{bmatrix}
        v_x\cos\theta \\
        v_x\sin\theta \\
        \omega \\
    \end{bmatrix}. 
\end{align}

Considering the computation cost of the integration to get the reference position $\boldsymbol p_i^r$ from the reference MS trajectory, the controller, upon receiving a trajectory, will perform pre-integration at a specific sampling interval $\overline t^{sp}$ and store the results $\{\boldsymbol p_j^{sp}\}$. 
When calculating the reference position $\boldsymbol p_i^r$, the controller will find the nearest pre-integrated time $j\overline t^{sp}$ and corresponding position $\boldsymbol p^{sp}_j$, and start the integral process from $\boldsymbol p^{sp}_j$ to get the more accurate value.

\section{Benchmark and Simulations} \label{sec:sim_exp}
% \todo{overview here}
In this section, we first evaluate the numerical integration errors caused by the trajectory representation, which are proved to have trivial effects. 
Then we validate the effectiveness of the proposed method and test it in various environments. 
We recommend readers refer to the submitted video for more detailed comparisons. 

\subsection{Integral Error Analysis}\label{sec:IntegralError}
\begin{figure}[t]
    \centering
    % \vspace{-0.3cm}
    \setlength{\abovecaptionskip}{1pt}
    \includegraphics[width=8.5cm]{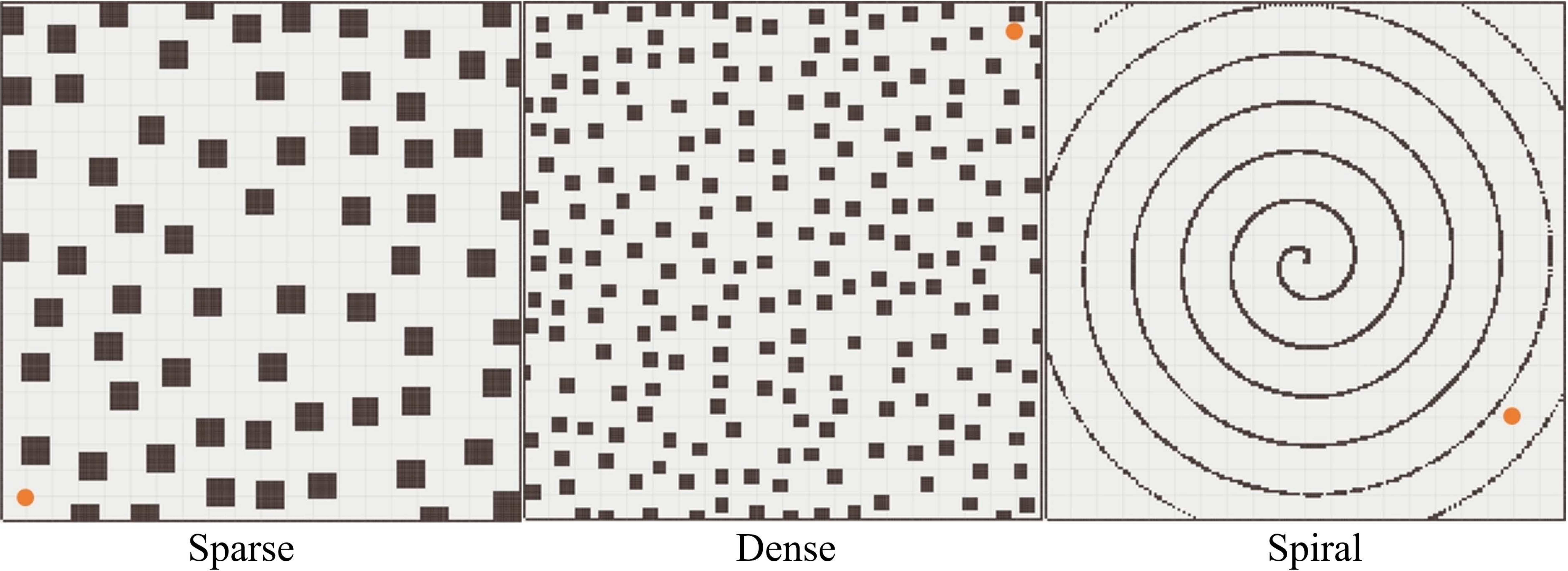}
    \caption{Integral error test environment. The orange point is the starting position. Final positions are randomly sampled and not plotted.}
    \label{fig:intergral_env}
    \vspace{-0.25cm}
\end{figure}

In this section, we focus on the impact of integration error. 
To test the general performance, we design three environments based on the difficulty of planning as shown in Fig.\ref{fig:intergral_env}:

\textbf{Sparse}: A $20\text{m} \times 20\text{m}$ environment with 65 randomly generated square obstacles with a size of $1\text{m} \times 1\text{m}$. 
\textbf{Dense}: A $20\text{m} \times 20\text{m}$ environment with 213 randomly generated square obstacles with a size of $0.5\text{m} \times 0.5\text{m}$.
The space between obstacles is significantly reduced, making navigation more challenging. 
\textbf{Spiral}: A spiral environment with size $20m\times20m$, where only one turning direction is allowed to move from the outside to the inside and the robot needs to rotate continuously and move over long distances.

The maximum velocity of the robot is set to $3\text{m/s}$, and the maximum angular velocity is set to $4\text{rad/s}$.
The initial value of trajectory time is set to $0.8\text{s}$ per segment, and the number of intervals $n$ per segment is set to $10$. 
We randomly sample the final positions within the reachable area of the map for more than $1000$ experiments for statistical generalization.

\begin{figure}[t]
    \centering
    \vspace{-0.2cm}
    \setlength{\abovecaptionskip}{3pt}
    \includegraphics[width=8.8cm]{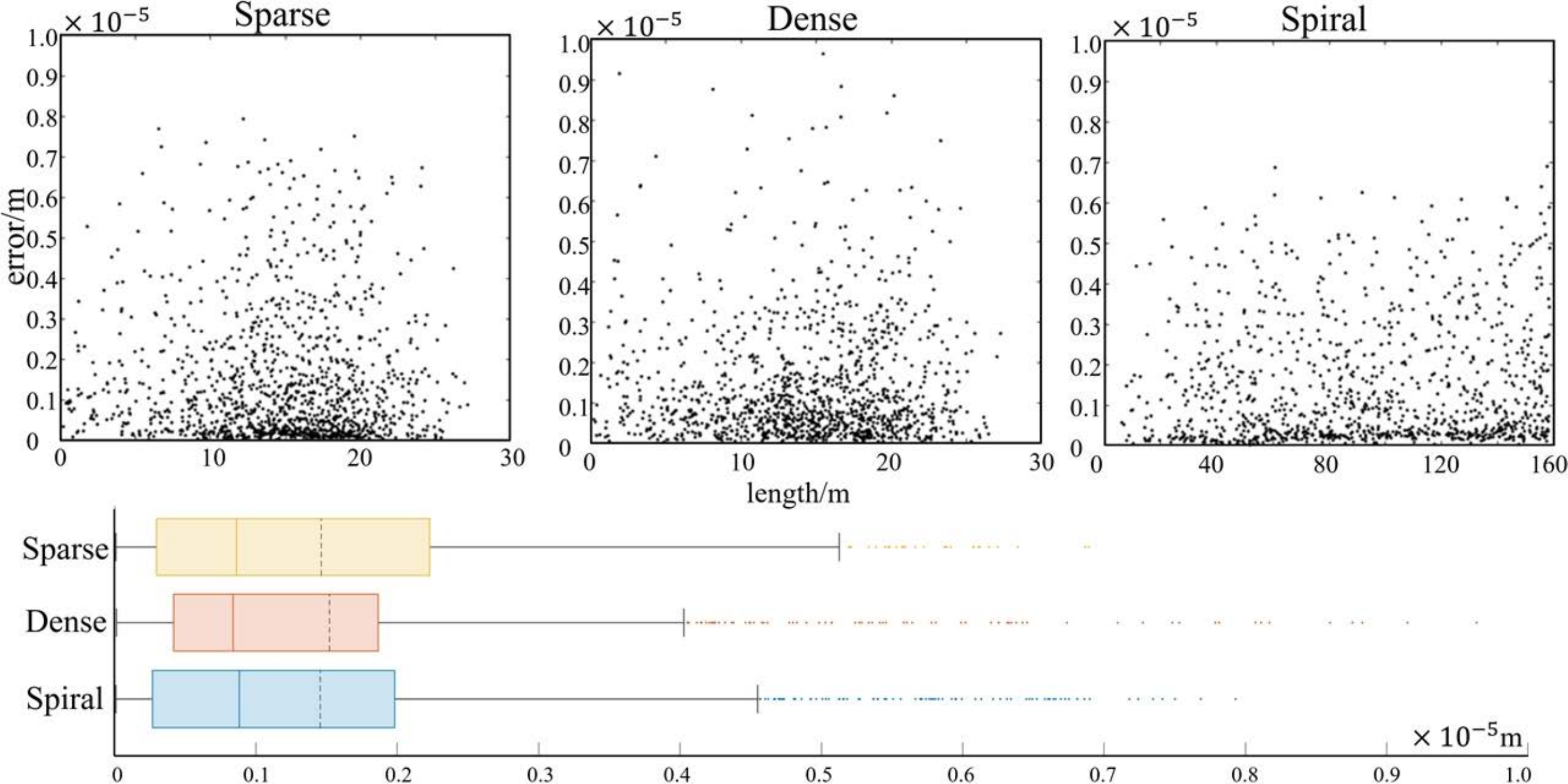}
    \caption{Integral error test results. 
            The x-axis of the top graph is the trajectory length and the y-axis is the final position error. 
            The bottom graph shows the statistical results, with outliers defined as values more than $1.5\text{IQR}$ (interquartile range) from the top of the box.}
    \label{fig:intergral_error}
    \vspace{-0.65cm}
\end{figure}

The results, as shown in Fig.\ref{fig:intergral_error}, indicate that the integration error is below $10^{-5}\text{m}$ for all maps and most lengths, which demonstrates that the integration error will not impact the performance of our proposed trajectory optimization. 
Therefore, it is reasonable to ignore the integration error. 

\subsection{Simulation and Benchmarks}
% We evaluated the proposed method in simulations to validate its effectiveness.

\begin{table*}[t]
    \centering
    \caption{Comparison of kinematics in different cases.}
    \vspace{-3pt}
    \label{tab:kine_sta}
    \begin{tabular}{c|c|cccc|cccc|cccc}
        \hline
        \multirow{2}{*}{\begin{tabular}[c]{@{}c@{}}Num \\ of \\ Obstacles\end{tabular}} & Length   & \multicolumn{4}{c|}{0-10m}                                                                                                                                                                                                            & \multicolumn{4}{c|}{10-20m}                                                                                                                                                  & \multicolumn{4}{c}{20m+}                                                                                                                                                                                                                                                                      \\ \cline{2-14} 
                                                                                        & Method   & {\begin{tabular}[c]{@{}c@{}}MLA\\ $m/s^2$\end{tabular}} & {\begin{tabular}[c]{@{}c@{}}MLJ\\ $m/s^3$\end{tabular}} & {\begin{tabular}[c]{@{}c@{}}MYA\\ $s^{-2}$\end{tabular}} & \begin{tabular}[c]{@{}c@{}}MYJ\\ $s^{-3}$\end{tabular} & {\begin{tabular}[c]{@{}c@{}}MLA\\ $m/s^2$\end{tabular}} & {\begin{tabular}[c]{@{}c@{}}MLJ\\ $m/s^3$\end{tabular}} & {\begin{tabular}[c]{@{}c@{}}MYA\\ $s^{-2}$\end{tabular}} & \begin{tabular}[c]{@{}c@{}}MYJ\\ $s^{-3}$\end{tabular} & {\begin{tabular}[c]{@{}c@{}}MLA\\ $m/s^2$\end{tabular}} & {\begin{tabular}[c]{@{}c@{}}MLJ\\ $m/s^3$\end{tabular}} & {\begin{tabular}[c]{@{}c@{}}MYA\\ $s^{-2}$\end{tabular}} & \begin{tabular}[c]{@{}c@{}}MYJ\\ $s^{-3}$\end{tabular} \\ \hline
        \multirow{3}{*}{50}                                                             & Proposed & {0.870}                                                 & {\textbf{1.215}}                                        & {\textbf{1.011}}                                         & \textbf{2.148}                                         & {\textbf{0.709}}                                        & {\textbf{0.913}}                                        & {\textbf{0.803}}                                         & \textbf{1.732}                                         & {\textbf{0.560}}                                        & {\textbf{0.750}}                                        & {\textbf{0.719}}                                                  & \textbf{1.536}                                         \\ 
                                                                                        & TEB      & {1.147}                                                 & {11.126}                                                 & {2.805}                                                  & 55.875                                                 & {0.898}                                                 & {8.820}                                                & {2.424}                                                  & 48.601                                                 & {0.763}                                                  & {7.421}                                                & {2.191}                                                  & 44.065                                                 \\  
                                                                                        & DF       & {\textbf{0.820}}                                        & {2.071}                                                 & {2.026}                                                  & 10.166                                                 & {0.864}                                                 & {2.034}                                                 & {1.909}                                                  & 9.812                                                  & {0.823}                                                 & {2.080}                                                 & {2.038}                                                  & 10.248                                                 \\ \hline
        \multirow{3}{*}{100}                                                            & Proposed & {0.806}                                                 & {\textbf{1.115}}                                        & {\textbf{1.084}}                                         & \textbf{2.353}                                         & {\textbf{0.673}}                                                 & {\textbf{0.879}}                                        & {\textbf{0.904}}                                         & \textbf{2.029}                                         & {\textbf{0.554}}                                        & {\textbf{0.749}}                                        & {\textbf{0.881}}                                         & \textbf{1.994}                                         \\  
                                                                                        & TEB      & {1.259}                                                 & {12.132}                                                 & {3.153}                                                  & 62.832                                                 & {1.040}                                                 & {10.075}                                                 & {2.894}                                                  & 58.122                                                 & {0.915}                                                 & {8.777}                                                & {2.811}                                                  & 56.753                                                 \\  
                                                                                        & DF       & {\textbf{0.735}}                                        & {1.813}                                                 & {1.764}                                                  & 7.870                                                  & {0.735}                                                 & {1.664}                                                 & {1.468}                                                  & 7.019                                                  & {0.739}                                                 & {1.820}                                                 & {1.767}                                                  & 7.935                                                  \\ \hline
        \multirow{3}{*}{200}                                                            & Proposed & {0.686}                                                 & {\textbf{0.970}}                                        & {\textbf{1.152}}                                         & \textbf{2.572}                                         & {\textbf{0.591}}                                        & {\textbf{0.801}}                                        & {\textbf{1.067}}                                         & \textbf{2.467}                                         & {\textbf{0.512}}                                        & {\textbf{0.697}}                                        & {\textbf{1.076}}                                         & \textbf{2.520}                                         \\  
                                                                                        & TEB      & {1.395}                                                 & {14.027}                                                 & {3.527}                                                  & 70.217                                                 & {1.261}                                                 & {13.347}                                                 & {3.429}                                                  & 68.574                                                 & {1.164}                                                 & {12.108}                                                 & {3.334}                                                  & 66.916                                                 \\  
                                                                                        & DF       & {\textbf{0.673}}                                        & {1.662}                                                 & {1.655}                                                  & 6.934                                                  & {0.643}                                                 & {1.490}                                                 & {1.343}                                                  & 5.915                                                  & {0.672}                                                 & {1.662}                                                 & {1.656}                                                  & 6.925                                                  \\ \hline
        \end{tabular}
    \vspace{-3pt}
\end{table*}

\begin{table*}[t]
    \centering
    \vspace{-3pt}
    \caption{Comparison of computation time, trajectory information and success rate in different cases.}
    \vspace{-3pt}
    \label{tab:other_sta}
    \setlength{\tabcolsep}{5pt}
    \begin{tabular}{c|c|ccccc|ccccc|ccccc}
        \hline
        \multirow{2}{*}{\begin{tabular}[c]{@{}c@{}}Num\\ of\\ Obstacles\end{tabular}} & Length   & \multicolumn{5}{c|}{0-10m}                                                                                                                                                                                                                                   & \multicolumn{5}{c|}{10-20m}                                                                                                                                                                                                                                      & \multicolumn{5}{c}{20m+}                                                                                                                                                                                  \\ \cline{2-17} 
                                                                                      & Method   & \begin{tabular}[c]{@{}c@{}}CT\\ $s$ \end{tabular} & \begin{tabular}[c]{@{}c@{}}TL\\ $m$\end{tabular} & \begin{tabular}[c]{@{}c@{}}TD\\ $s$\end{tabular} & \begin{tabular}[c]{@{}c@{}}MV\\ $m/s$\end{tabular} & \begin{tabular}[c]{@{}c@{}}SR\\ $\%$\end{tabular} & \begin{tabular}[c]{@{}c@{}}CT\\ $s$ \end{tabular} & \begin{tabular}[c]{@{}c@{}}TL\\ $m$\end{tabular} & \begin{tabular}[c]{@{}c@{}}TD\\ $s$\end{tabular}  & \begin{tabular}[c]{@{}c@{}}MV\\ $m/s$\end{tabular} & \begin{tabular}[c]{@{}c@{}}SR\\ $\%$\end{tabular} & \begin{tabular}[c]{@{}c@{}}CT\\ $s$ \end{tabular} & \begin{tabular}[c]{@{}c@{}}TL\\ $m$\end{tabular} & \begin{tabular}[c]{@{}c@{}}TD\\ $s$\end{tabular} & \begin{tabular}[c]{@{}c@{}}MV\\ $m/s$\end{tabular} & \begin{tabular}[c]{@{}c@{}}SR\\ $\%$\end{tabular} \\ \hline
        \multirow{3}{*}{50}                                                           & Proposed & \textbf{0.011}                                   & 6.822                                           & 5.726                                            & 1.192                                               & \textbf{100}                                      & \textbf{0.024}                                   & 14.73                                           & 8.330                                            & 1.768                                                & \textbf{100}                                      & \textbf{0.045}                                   & 22.48                                           & 11.28                                            & 1.993                                            & \textbf{100}                                      \\
                                                                                      & TEB      & 0.271                                            & 6.409                                           & 5.676                                            & 1.129                                               & \textbf{100}                                      & 0.557                                            & 14.52                                           & 10.61                                            & 1.369                                                & \textbf{100}                                      & 0.814                                            & 22.25                                           & 15.29                                            & 1.455                                            & \textbf{100}                                             \\
                                                                                      & DF       & 0.060                                            & 7.474                                           & 6.566                                            & 1.139                                               & 82.02                                             & 0.081                                            & 15.39                                           & 9.905                                            & 1.554                                                & 86.11                                             & 0.100                                            & 22.65                                           & 12.86                                            & 1.761                                            & 88.45                                             \\ \hline
        \multirow{3}{*}{100}                                                          & Proposed & \textbf{0.011}                                   & 6.801                                           & 6.094                                            & 1.116                                               & \textbf{100}                                      & \textbf{0.026}                                   & 14.70                                           & 9.010                                            & 1.632                                                & \textbf{100}                                      & \textbf{0.054}                                   & 22.41                                           & 12.19                                            & 1.838                                            & \textbf{100}                                      \\
                                                                                      & TEB      & 0.398                                            & 6.430                                           & 5.754                                            & 1.117                                               & 99.92                                             & 1.028                                            & 14.56                                           & 10.73                                            & 1.357                                                & \textbf{100}                                      & 1.624                                            & 22.18                                           & 15.30                                            & 1.450                                            & \textbf{100}                                             \\
                                                                                      & DF       & 0.062                                            & 6.953                                           & 6.055                                            & 1.148                                               & 83.23                                             & 0.082                                            & 14.81                                           & 9.026                                            & 1.641                                                & 89.51                                             & 0.098                                            & 22.34                                           & 11.91                                            & 1.876                                            & 86.63                                             \\ \hline
        \multirow{3}{*}{200}                                                          & Proposed & \textbf{0.012}                                   & 6.874                                           & 6.800                                            & 1.011                                               & \textbf{99.52}                                    & \textbf{0.033}                                   & 14.75                                           & 10.47                                            & 1.409                                                & \textbf{98.93}                                    & \textbf{0.068}                                   & 22.80                                           & 14.46                                            & 1.577                                            & \textbf{99.84}                                    \\
                                                                                      & TEB      & 0.768                                            & 6.392                                           & 6.197                                            & 1.031                                               & 94.57                                             & 2.110                                            & 14.37                                           & 11.42                                            & 1.258                                                & 88.95                                             & 3.776                                            & 22.31                                           & 16.73                                            & 1.334                                            & 83.71                                             \\
                                                                                      & DF       & 0.059                                            & 7.142                                           & 6.413                                            & 1.114                                               & 81.51                                             & 0.079                                            & 14.88                                           & 9.687                                            & 1.536                                                & 86.72                                             & 0.099                                            & 22.85                                           & 12.89                                            & 1.773                                            & 88.09                                             \\ \hline
    \end{tabular}
    \vspace{-0.6cm}
\end{table*}

Experimental Setup: All simulation experiments are conducted on a desktop computer running Ubuntu 18.04, equipped with an Intel Core i7-10700K CPU. 
We compared our proposed method with two widely recognized motion planning algorithms for DD robots: Timed Elastic Band (TEB) planner \cite{rosmann2013efficient} and a Differential Flatness-based (DF) planner \cite{zhang2023trajectory}. 
TEB considers temporal optimization while guaranteeing path continuity and smoothing through an elastic band model.
Our reference for the DF method is drawn from our prior work, which notably offers a more constrained approach for angular velocity and acceleration, crucial for adhering to the robot's kinematic constraints because of nonholonomic dynamics. 
All methods utilize a global pointcloud for map construction and set the kinematic model to a SDD model.
Considering robot's decoupled velocity and angular acceleration constraints, we set lower bounds for TEB to maintain kinematic feasibility. 
For the DF method, which requires pre-specification of the robot's movement direction (forward or backward), we simplify this by assuming only forward movement. 
We model the robot as a point mass and set the same safety distance for all methods.

Quantitative Evaluation: We perform extensive quantitative tests in environments of size $20m\times 20m$ with randomly generated obstacles. 
We use different numbers of obstacles to characterize the complexity of the environment, while categorizing the length of trajectories according to the distance between the starting and final positions. 
In each case, we randomly assign feasible starting and ending positions and conduct at least 1000 experiments.
In each experiment, we record the kinematics of the trajectories, including the mean linear acceleration (MLA), the mean linear jerk (MLJ), the mean angular acceleration (MYA), and the mean angular jerk (MYJ), as shown in Table.\ref{tab:kine_sta}. 
We also compute the average computation time (CT), the trajectory length (TL), the trajectory duration (TD), the mean velocity (MV) and the success rate (SR) for each case, as shown in Table.\ref{tab:other_sta}. 
The optimization is considered to fail when it does not converge or penalty functions are not satisfied. 

\begin{figure}[t]
    \centering
    % \vspace{-0.6cm}
    \setlength{\abovecaptionskip}{4pt}
    \includegraphics[width=8.8cm]{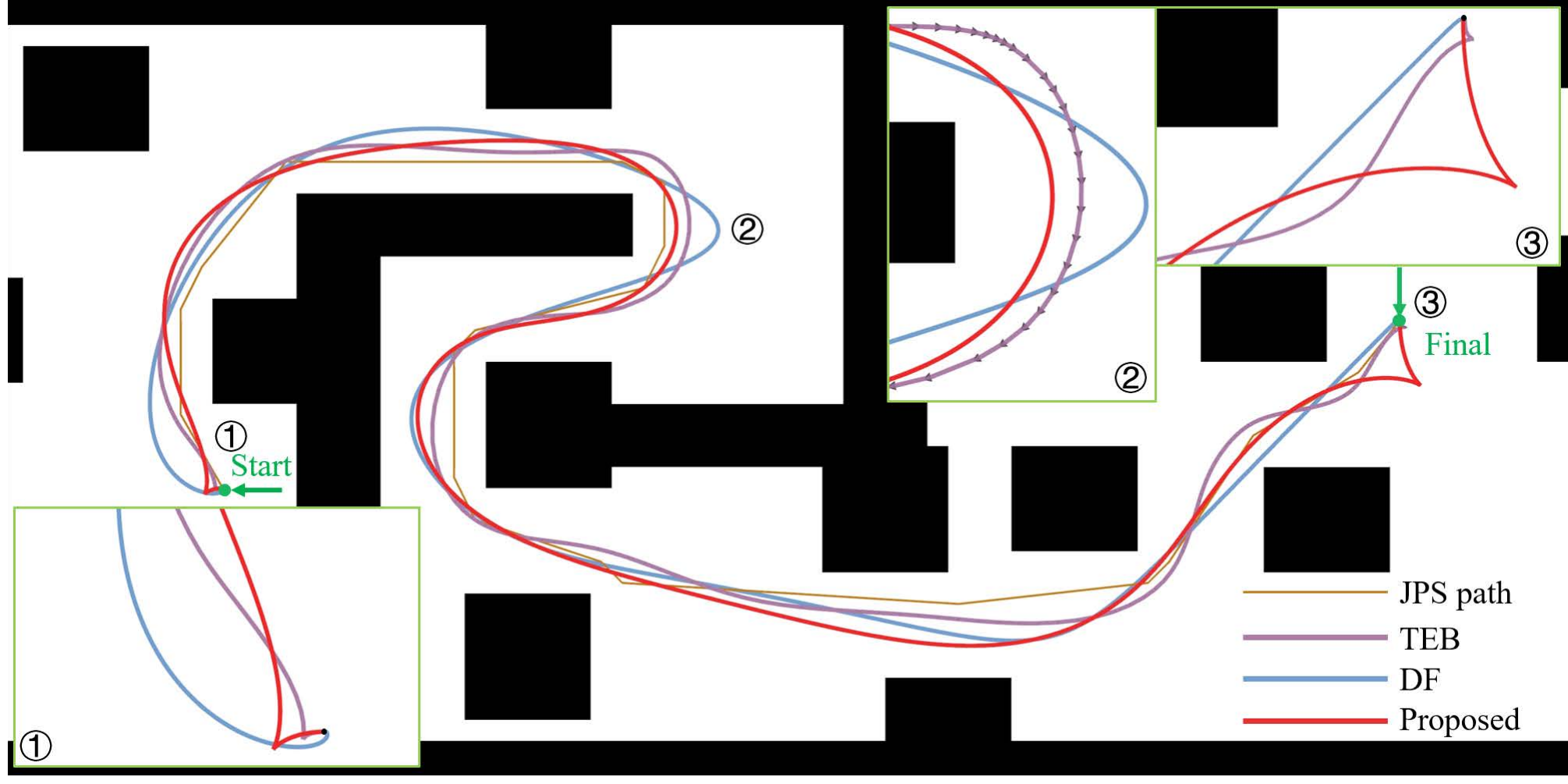}
    \caption{Visual comparison of different methods. Proposed method and TEB are set to allow moving backward.
    To better illustrate the details, we zoom in three areas(\textcircled{1} to \textcircled{3}) with distinct differences.
    Specifically, in region \textcircled{2}, we show the via-points of TEB. }
    \label{fig:comp_method}
    \vspace{-0.65cm}
\end{figure}

One of the simulation results is shown in Fig.\ref{fig:comp_method}. 
Simulation results suggest that the proposed method notably outperforms others in trajectory performance.
TEB generates trajectories composed of via-points, which can only generate piecewise smooth trajectories.
As shown in Table.\ref{tab:kine_sta}, kinematic metrics of the proposed method demonstrate superior smoothness. 
By modeling the trajectory as a high-order polynomial, we ensure the continuity, making the trajectory easier for the controller to track compared to the TEB method with discrete control input quantities. 
Due to the nonholonomic dynamics, although \cite{zhang2023trajectory} proposed methods to make the DF trajectory better satisfy the kinematic constraints, these methods are limited to being "confined within certain limits", such as the angular acceleration.
As shown in Table.\ref{tab:kine_sta}, the values of MYA and MYJ are both larger than those of the proposed method, which will lead to overly aggressive motion and negatively affect the control of the orientation angle.
The proposed method can adapt to different scenarios and is robust under various environmental complexities and trajectory lengths. 
In contrast, the nonlinearity of DF affects the computational efficiency and still fails to satisfy constraints such as angular acceleration in some scenarios. 
TEB requires more control points by discretizing trajectories, which will lead to longer computation time or more difficult convergence.

\begin{figure}[b]
    \centering
    \vspace{-0.7cm}
    \setlength{\abovecaptionskip}{4pt}
    \includegraphics[width=8.8cm]{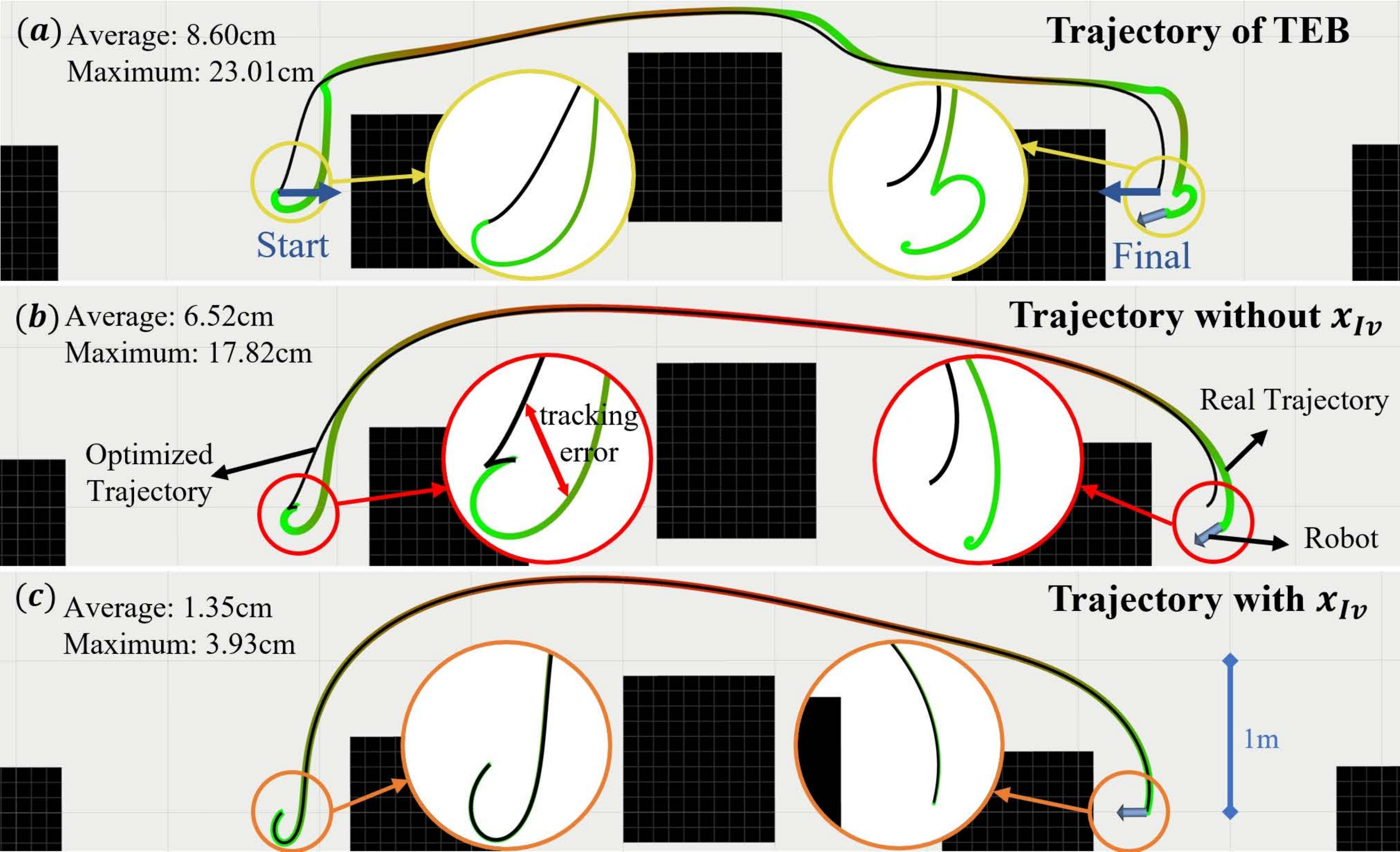}
    \caption{Planning results of TEB (a), and the proposed method without (b) and with (c) lateral slip of $x_{Iv}$, when lateral slip cannot be ignored. 
    We calculate the average tracking error and maximum tracking error of the controller. 
    By modeling lateral slip into the MS trajectory, the controller achieves better performance in tracking the proposed trajectory.
    }
    \label{fig:tracked_traj}
    % \vspace{-0.65cm}
\end{figure}

When considering the case of $x_{Iv} \neq 0$, which is common in TDD robots, the interaction between the tracks and the ground causes lateral movement of the geometric center. 
To validate the effectiveness of the proposed method, we set $y_{Il} = -y_{Ir} = 0.3m$ and $x_{Iv} = 0.2m$ under the same environmental settings as before..
We recorded the computation time ($CT$), the mean velocity of the geometric center ($MV$), and the success rate ($SR$), as shown in Table.\ref{tab:icr}. 
The proposed method consistently achieved feasible trajectories with high efficiency and robustness across various environments. 
However, the increased complexity of the kinematic model resulted in a slight increase in computation time. 
The average velocity of the geometric center also increased, which is mainly caused by the lateral slip.
Due to the lateral slip, the geometric center cannot rotate with zero radius, resulting in poor trajectory tracking if $x_{Iv}$ is not considered, as illustrated in Fig.\ref{fig:tracked_traj}(b). 
TEB plans the lateral velocity $v_y$ separately; however, in reality, $v_y$ is related to $\omega$ as described in Eq.\ref{ali:vy}. 
Independently planning $v_y$ does not conform to the robot's motion model, leading to significant tracking errors, as shown in Fig.\ref{fig:tracked_traj}(a). 
In contrast, the proposed method incorporates slip into the trajectory representation, resulting in improved tracking performance, as demonstrated in Fig.\ref{fig:tracked_traj}(c).

\begin{table}[h]
    \centering
    \vspace{-6pt}
    \caption{Comparison when considering the case of $x_{Iv} \neq 0$.}
    \vspace{-3pt}
    \label{tab:icr}
    \setlength{\tabcolsep}{2.5pt}
    \begin{tabular}{c|ccc|ccc|ccc}
        \hline
        \multirow{2}{*}{\begin{tabular}[c]{@{}c@{}}Num\\ of\\ Obstacles\end{tabular}} & \multicolumn{3}{c|}{0-10m}                                                                                                                                 & \multicolumn{3}{c|}{10-20m}                                                                                                                                & \multicolumn{3}{c}{20m+}                                                                                                                                   \\ \cline{2-10} 
                                                                                      & \begin{tabular}[c]{@{}c@{}}CT\\ $ms$\end{tabular} & \begin{tabular}[c]{@{}c@{}}MV\\ $m/s$\end{tabular} & \begin{tabular}[c]{@{}c@{}}SR\\ $\%$\end{tabular} & \begin{tabular}[c]{@{}c@{}}CT\\ $ms$\end{tabular} & \begin{tabular}[c]{@{}c@{}}MV\\ $m/s$\end{tabular} & \begin{tabular}[c]{@{}c@{}}SR\\ $\%$\end{tabular} & \begin{tabular}[c]{@{}c@{}}CT\\ $ms$\end{tabular} & \begin{tabular}[c]{@{}c@{}}MV\\ $m/s$\end{tabular} & \begin{tabular}[c]{@{}c@{}}SR\\ $\%$\end{tabular} \\ \hline
        50                                                                            & 14.54                                             & 1.335                                              & 100                                               & 15.97                                             & 1.268                                              & 99.84                                             & 18.48                                             & 1.179                                              & 99.91                                             \\
        100                                                                           & 40.57                                             & 1.824                                              & 99.91                                             & 44.29                                             & 1.712                                              & 100                                               & 46.83                                             & 1.513                                              & 99.61                                             \\
        200                                                                           & 70.09                                             & 1.982                                              & 100                                               & 76.18                                             & 1.822                                              & 100                                               & 75.61                                             & 1.312                                              & 99.43                                             \\ \hline
    \end{tabular}
    \vspace{-6pt}
\end{table}

\begin{figure}[h]
    \centering
    \vspace{-0.3cm}
    \setlength{\abovecaptionskip}{4pt}
    \includegraphics[width=8.8cm]{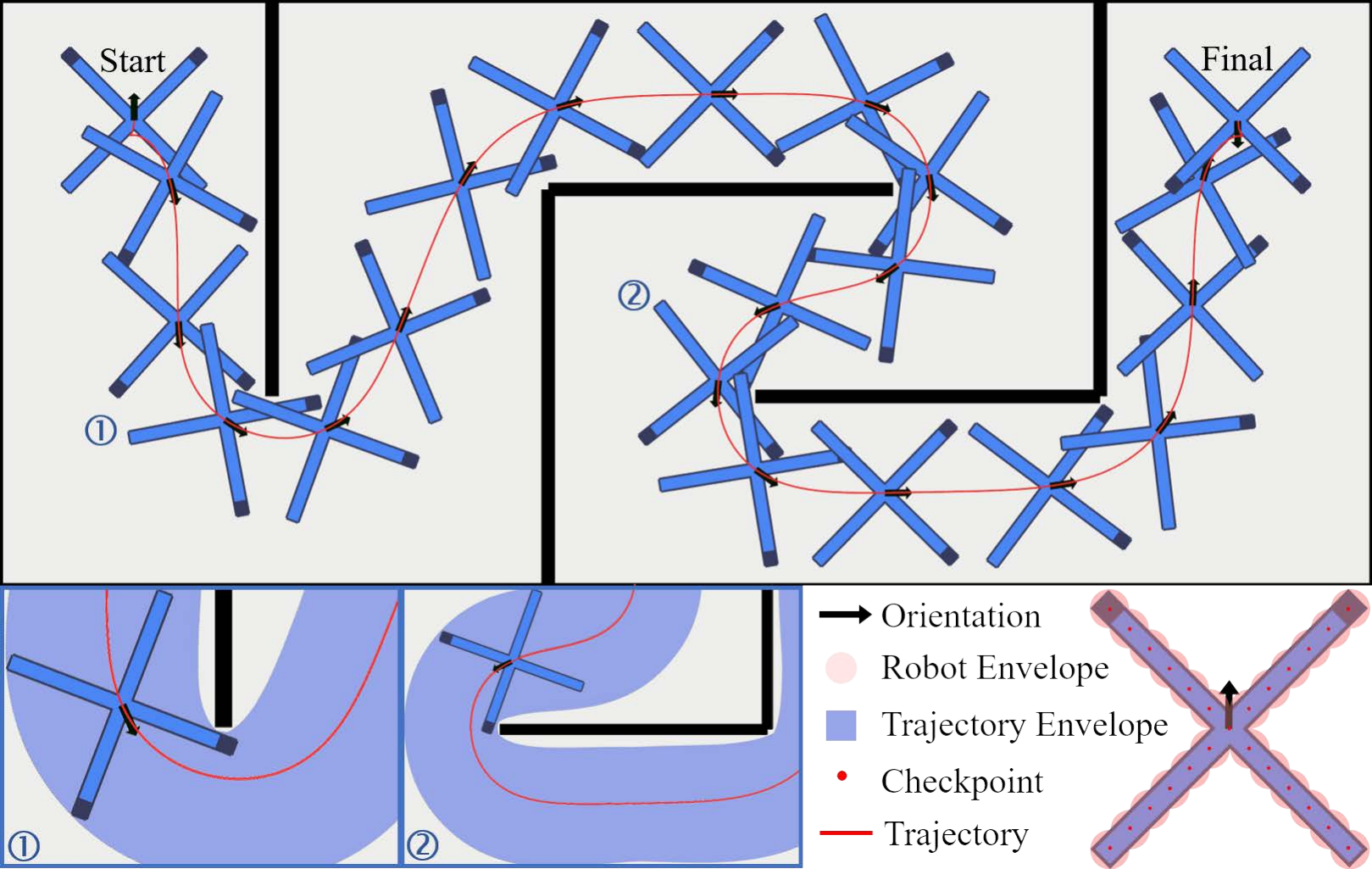}
    \caption{Trajectory optimization considering the robot's shape. 
    The robot is X-shaped designed, with selected checkpoints marked on it shown in the bottom right. 
    Circles centered on these checkpoints are used to enclose the robot's outline. 
    Two parts of the trajectory(\textcircled{1} and \textcircled{2}) are zoomed in to illustrate the trajectory envelope in the bottom left,  showing how the optimized trajectory considers the robot shapes.    
    }
    \label{fig:non_particle}
    \vspace{-0.6cm}
\end{figure}

When the shape cannot be simply modeled as a circle, we can select checkpoints along the robot's contour to better model its shape. 
As shown in Fig.\ref{fig:non_particle}, using safety circles centered at these checkpoints can construct the robot's envelope with as small as possible error.
By applying the safety constraints in Sec.\ref{subsec:safety}, the proposed method can fully utilize the safe space while ensuring the safety of the trajectory.

\subsection{Replanning Experiments}\label{subsec:replan}

\begin{figure}[t]
    \centering
    % \vspace{-0.5cm}
    \setlength{\abovecaptionskip}{4pt}
    \includegraphics[width=8.8cm]{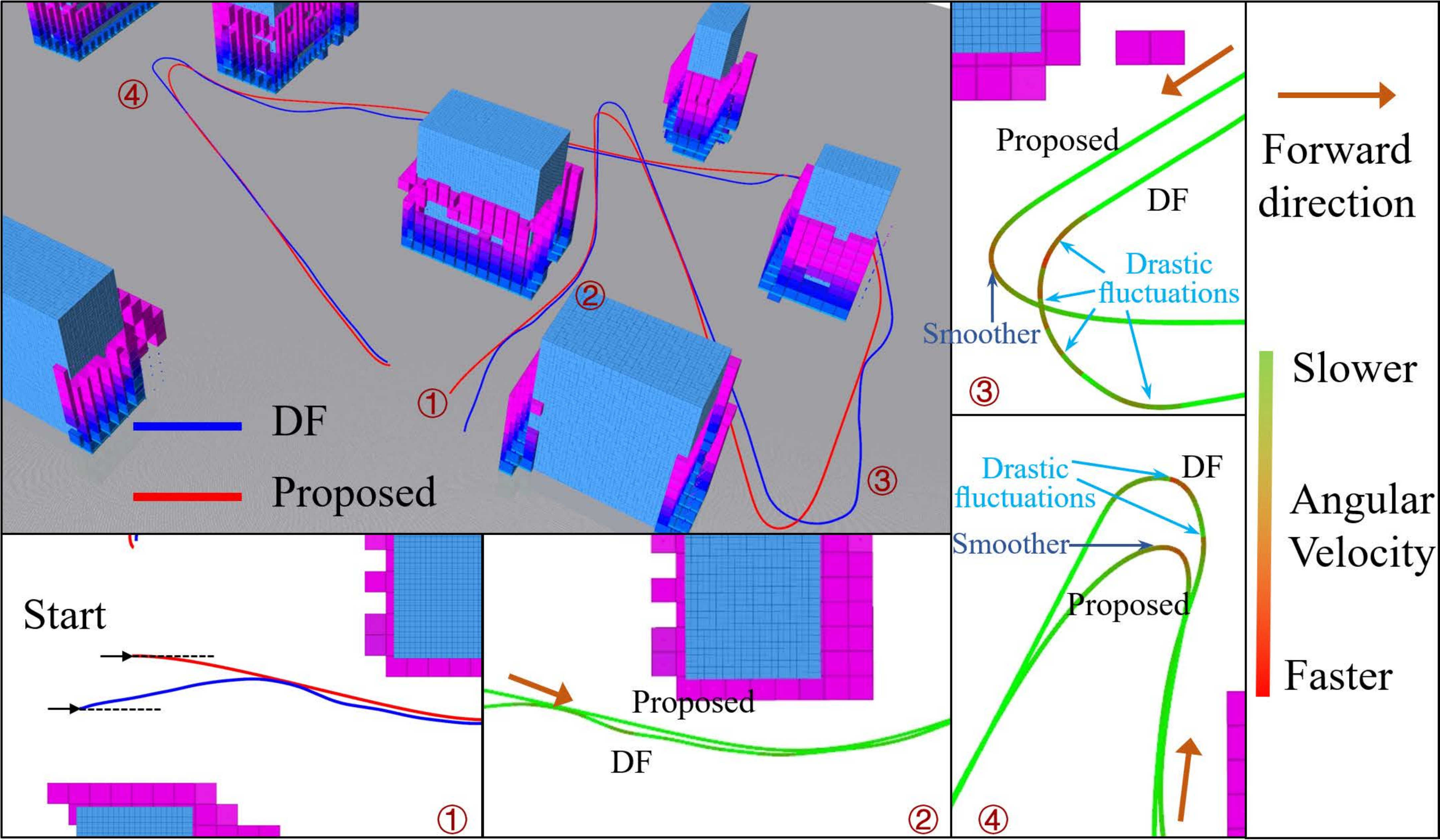}
    % \captionsetup{skip=0pt}
    \caption{Patrolling with DF and Proposed methods, the next target is provided to the planner upon nearing the current target.
            We zoom in four position(\textcircled{1} -- \textcircled{4}) to show the smoothness and efficiency difference.
            \textcircled{1} is the starting position, with the initial state directed towards the positive right-hand side. 
            We use the gradient color to show the angular velocity in \textcircled{2}-\textcircled{4}. 
            The robot moves around the obstacle in \textcircled{2}. 
            In \textcircled{3} and \textcircled{4}, a new target is provided to the planner, prompting the robot to turn towards the new target.
    }
    \label{fig:patrol}
    \vspace{-0.6cm}
\end{figure}

Additional tests are conducted for the two methods: the DF method and the proposed method, both of which fulfill real-time requirements.
A series of targets are set up within the environment, and the robot should sequentially reach these targets, as shown in Fig.\ref{fig:patrol}. 
Both methods replan at the same frequency. 
For clarity, the starting positions of the two methods are staggered, as indicated by Fig.\ref{fig:patrol}\textcircled{1}. 
The DF method uses differential flatness to represent the trajectory, where the robot's state can be expressed through Cartesian coordinates $x,y$ and their derivatives. 
Based on this representation, we can get the angular velocity $\omega$ and the angular acceleration $\alpha$:
\begin{equation}
\begin{aligned}[c]
    \omega & = \frac{\dot x\ddot y-\dot y\ddot x}{\dot x^2+\dot y^2},     \label{align:omegaalphaDF} \\
    \alpha & = \frac{\dot x \dddot y-\dot y \dddot x}{{\dot x^2+\dot y^2}}-\frac{2(\dot x\ddot y-\dot y\ddot x)(\dot x\ddot y+\dot y\ddot x)}{({\dot x^2+\dot y^2})^2}. 
\end{aligned}
\end{equation}
It is noteworthy that the linear velocity of the robot, $v_x = \sqrt{\dot x^2 + \dot y^2}$, in the denominator can adversely impact the performance of the DF method, particularly at low velocities.
When the robot starts moving, as shown in Fig.\ref{fig:patrol}\textcircled{1}, abrupt changes in orientation may occur within the DF trajectory. 
Furthermore, the nonlinearity of Eq.(\ref{align:omegaalphaDF}) affects the optimality of the DF trajectory, leading to discrepancies between pre- and post-replanning trajectories and compromising the smoothness of execution results. 
In contrast, the proposed method is devoid of such nonlinearity, yielding better execution results.
Given that the optimization objective Eq.(\ref{ali:minJ}) of the proposed method considers $\omega$ and $\alpha$, in contrast to the DF method which employs Eq.(\ref{align:omegaalphaDF}) as constraints without integrating it into minimizing control effort, the resulting angular velocity profile of the proposed trajectory exhibits a significantly smoother behavior, as demonstrated in Fig.\ref{fig:patrol}\textcircled{2}.
This difference becomes particularly pronounced when the robot is in motion and the target is behind it, as shown in Fig.\ref{fig:patrol}\textcircled{3} and \textcircled{4}. 
% In these scenarios, the DF trajectory experiences drastic angular velocity fluctuations during turns, adversely affecting control stability. 
% Conversely, the proposed method smoothly transitions from deceleration to turning, clearly demonstrating its novelty.

\begin{figure}[t]
    \centering
    % \vspace{-0.65cm}
    \includegraphics[width=8.8cm]{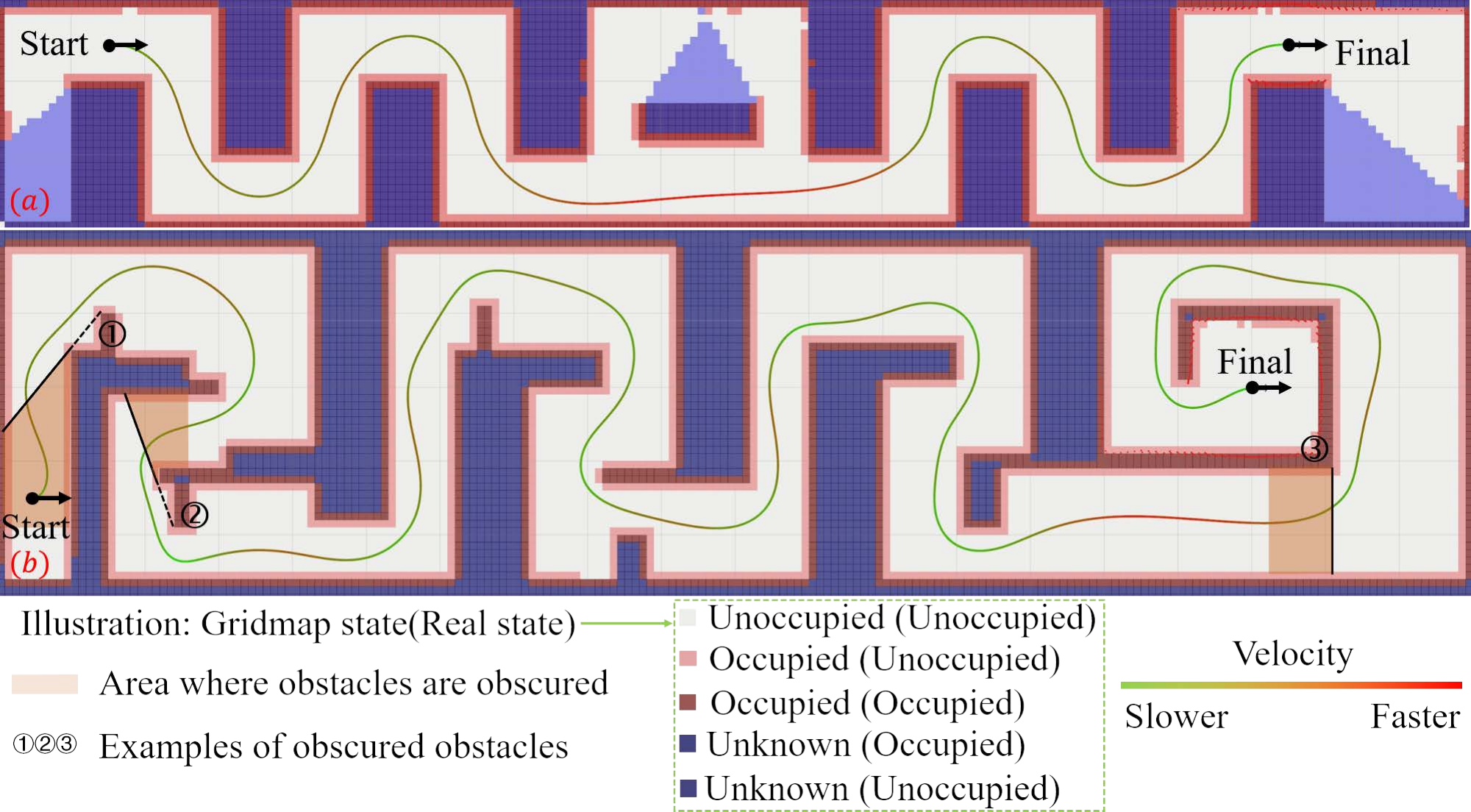}
    \setlength{\abovecaptionskip}{-12pt}
    % \captionsetup{skip=10pt}
    \caption{Simulation results for replanning. (a) A U-shaped environment. (b) A designed environment containing many obstacles that may be obscured.
    % \todo{caption in the fig is too small}
    }
    \label{fig:replan_env}
    \vspace{-0.6cm}
\end{figure}

In order to show the application of the proposed method in a fully autonomous simulation, we conduct replanning simulation tests in unknown maps, as shown in Fig.\ref{fig:replan_env}.
The simulator receives and executes velocity commands and broadcasts the current position of the robot. 
We simulate the Lidar using a laser simulator\footnote{\url{https://github.com/ZJU-FAST-Lab/laser_simulator}}.
The robot need to receive point clouds during runtime to build maps. 
The detection range is set to $7\text{m}$, the truncation length is $l_{max}=8\text{m}$. 
The convergence condition for Eq.(\ref{align:finalPos_emax}) is set to $e_{max}'=0.1\text{m}$, whereas without truncation, the convergence condition is $e_{max}=0.01\text{m}$.
The grid size of the map is $0.1\text{m}$ and the replanning frequency is set to $12.5\text{Hz}$.

We design a complex U-shaped environment, and the results are shown in Fig.\ref{fig:replan_env}(a). 
We tabulate the running time of each part, as shown in Table.\ref{tab:replan_time}, which proves that the proposed method can complete the computation in real-time. 
Here we set the expected computation time $T_R=20\text{ms}$, and in almost all cases, the planner can complete the computation in time. 
The expected computation time $T_R$ is much less than time used for mapping and actuator control, thereby minimizing the impact of the planner on the system's response speed and enhancing the robot's applicability in time-sensitive scenarios.

\begin{table}[h]
    \vspace{-0.25cm}
    \centering
    \caption{Average computation time of replanning.}
    \vspace{-3.5pt}
    \begin{tabular}{c|cccc}
        \hline
                                                                  & JPS   & Pre-process & Optimization & All   \\ \hline
        \begin{tabular}[c]{@{}c@{}}Comp. Time\\ (ms)\end{tabular} & 0.437 & 0.814       & 9.219        & 10.47 \\ \hline
    \end{tabular}
    \label{tab:replan_time}
    \vspace{-0.2cm}
\end{table}

To validate the effectiveness of the proposed method, we specifically design an environment as shown in Fig.\ref{fig:replan_env}(b), where some obstacles are easily obscured and rapid replanning is necessary to ensure safety.
We adopt the strategy from \cite{zhou2021raptor}, which involves forcing the robot to move away from obstacles to broaden the field of view. 
Our proposed method enables the robot to reach the final position safely.

\section{Real-world Experiments} \label{sec:real_exp}
To verify the practical application and universality of the proposed method, we conducted tests with different DD platforms in various environments:
\begin{itemize}
    \item A two-wheeled SDD robot with a pre-built map and the motion capture system, simulating tasks in flat indoor scenarios such as cleaning or service robots.
    \item A faster and more flexible four-wheeled SKDD robot equipped with Lidar for localization and mapping in various unknown environments, testing the method's ability to replan in complex unknown environments.
    \item A TDD robot equipped with Lidar for localization and mapping, evaluating the method's adaptability in narrow and unknown environments with low acceleration for common heavy-load tracked robots.
\end{itemize}

\subsection{Two-wheeled SDD Robots}

\begin{figure*}[h]
    \centering
    % \vspace{-0.6cm}
    \setlength{\abovecaptionskip}{1pt}
    \includegraphics[width=18cm]{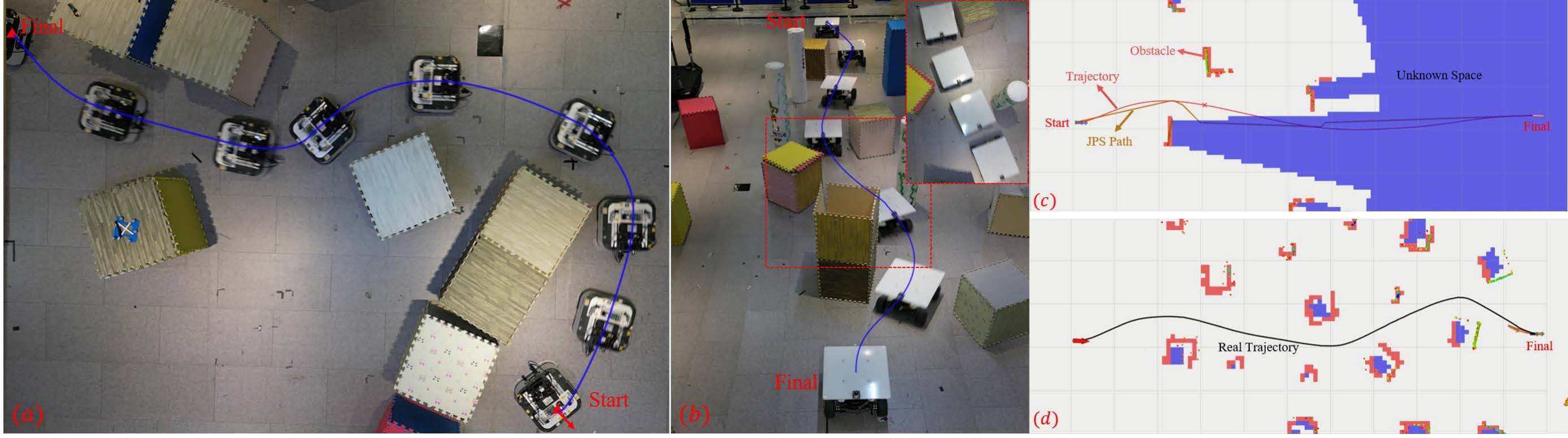}
    \caption{Planning results. (a)The two-wheeled SDD robot. (b)The four-wheeled SKDD robot. (c)Map and planning result at the starting point of the SKDD robot. (d)Real trajectory of the SKDD robot.}
    \label{fig:yunjing_xinwei}
    \setlength{\belowcaptionskip}{-0.4cm}
    \vspace{-0.7cm}
\end{figure*}

We employ the TRACER MINI\footnote{\url{https://www.agilex.ai/chassis/2}} as the two-wheeled SDD robot platform, which is controlled through linear and angular velocity.
We pre-build the map and utilize the motion capture system\footnote{\url{https://www.nokov.com/}} for localization. 
We intentionally task the robot with large-angle maneuvers, where there is a significant difference between the initial orientation and the upcoming forward direction.
We constrain the maximum velocity to $v_{\text{max}}\text{=}1.0\text{m/s}$ and the maximum angular velocity to $\omega_{\text{max}}\text{=}1.0\text{rad/s}$.
The result is shown in Fig.\ref{fig:yunjing_xinwei}(a), demonstrating the capability of our proposed method to produce smooth motion. 

\subsection{Four-Wheeled SKDD Robots}\label{subsec:four_wheel}
\begin{figure}[tb]
    \centering
    % \vspace{0.0cm}
    \setlength{\abovecaptionskip}{3pt}
    \includegraphics[width=8.8cm]{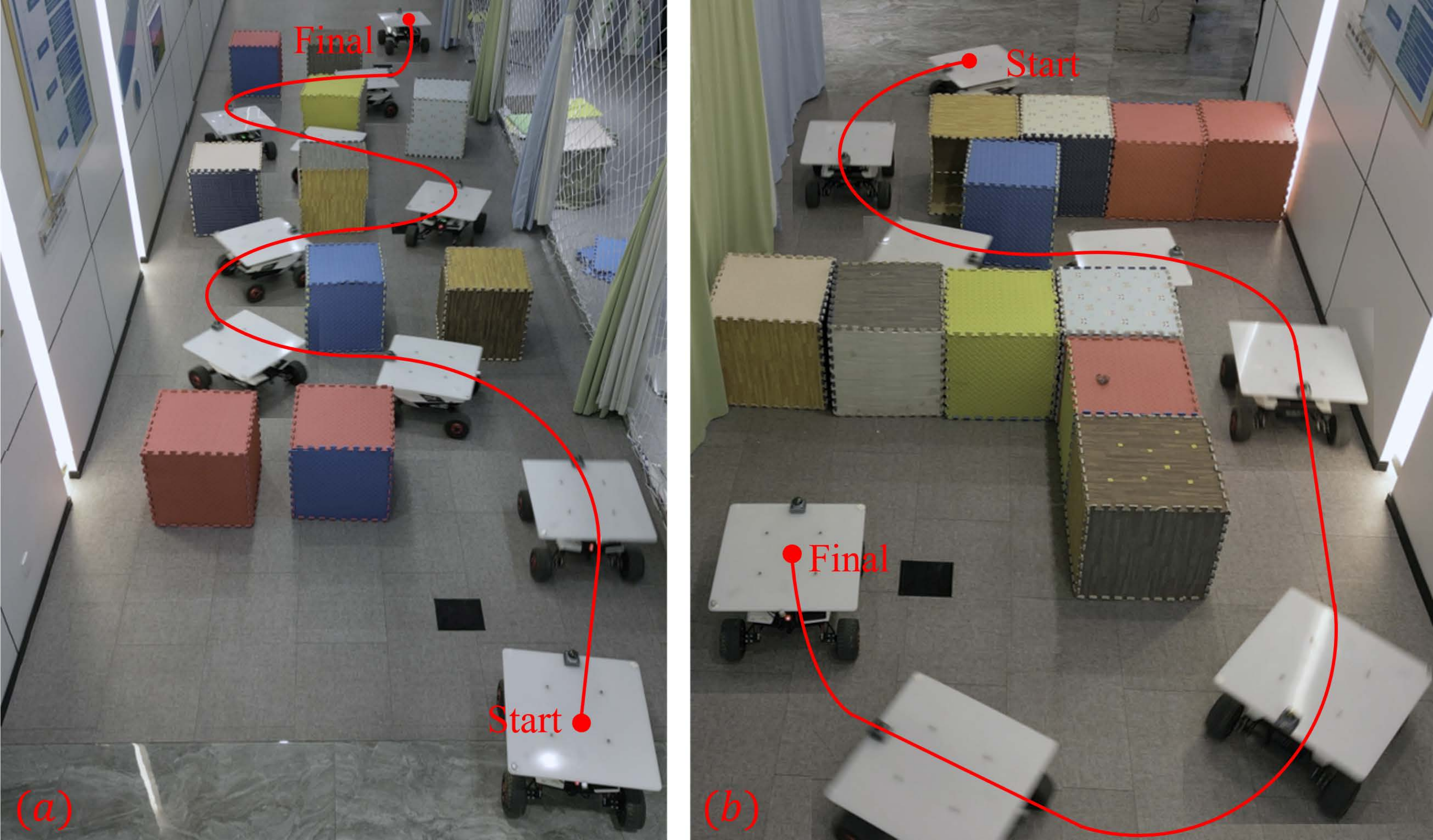}
    \caption{Planning results in narrow and complex spaces.}
    \label{fig:complex_exam}
    \vspace{-0.65cm}
\end{figure}

We employ the SCOUT MINI\footnote{\url{https://www.agilex.ai/chassis/11}} as our SKDD robot platform, which is controlled through linear and angular velocity. 
We randomly place obstacles and use FAST-LIO\cite{Xu2022FASTLIO2} for localization, and limit the perception range to $7\text{m}$.
The robot's maximum velocity is set to $v_{\text{max}}\text{=}2.0\text{m/s}$ and the maximum angular velocity to $\omega_{\text{max}}\text{=}1.5\text{rad/s}$.
Replanning is executed at a rate of $10.0\text{Hz}$, aligning with the frequency of FAST-LIO.
The combination of limited perception range, occluded obstacles, and high velocity presents a significant challenge to motion planning. 
Given the large running velocity and large robot size, the method must rapidly generate new trajectories to navigate sudden obstacles.
As shown in Fig.\ref{fig:yunjing_xinwei}(b), the robot travels through a random forest environment to reach the final position. 
At the starting position, due to obstruction, a part of the map is unknown, as shown in Fig.\ref{fig:yunjing_xinwei}(c). 
The robot can safely reach the final position while constructing the map, as shown in Fig.\ref{fig:yunjing_xinwei}(d).
In addition, we also specifically design narrow and complex maps to test the proposed method, as shown in Fig.\ref{fig:complex_exam}. 
Experimental results show that the proposed method can generate smooth motions within complex environments.

\subsection{TDD Robots}

\begin{figure}[tb]
    \centering
    % \vspace{0.0cm}
    \setlength{\abovecaptionskip}{3pt}
    \includegraphics[width=8.8cm]{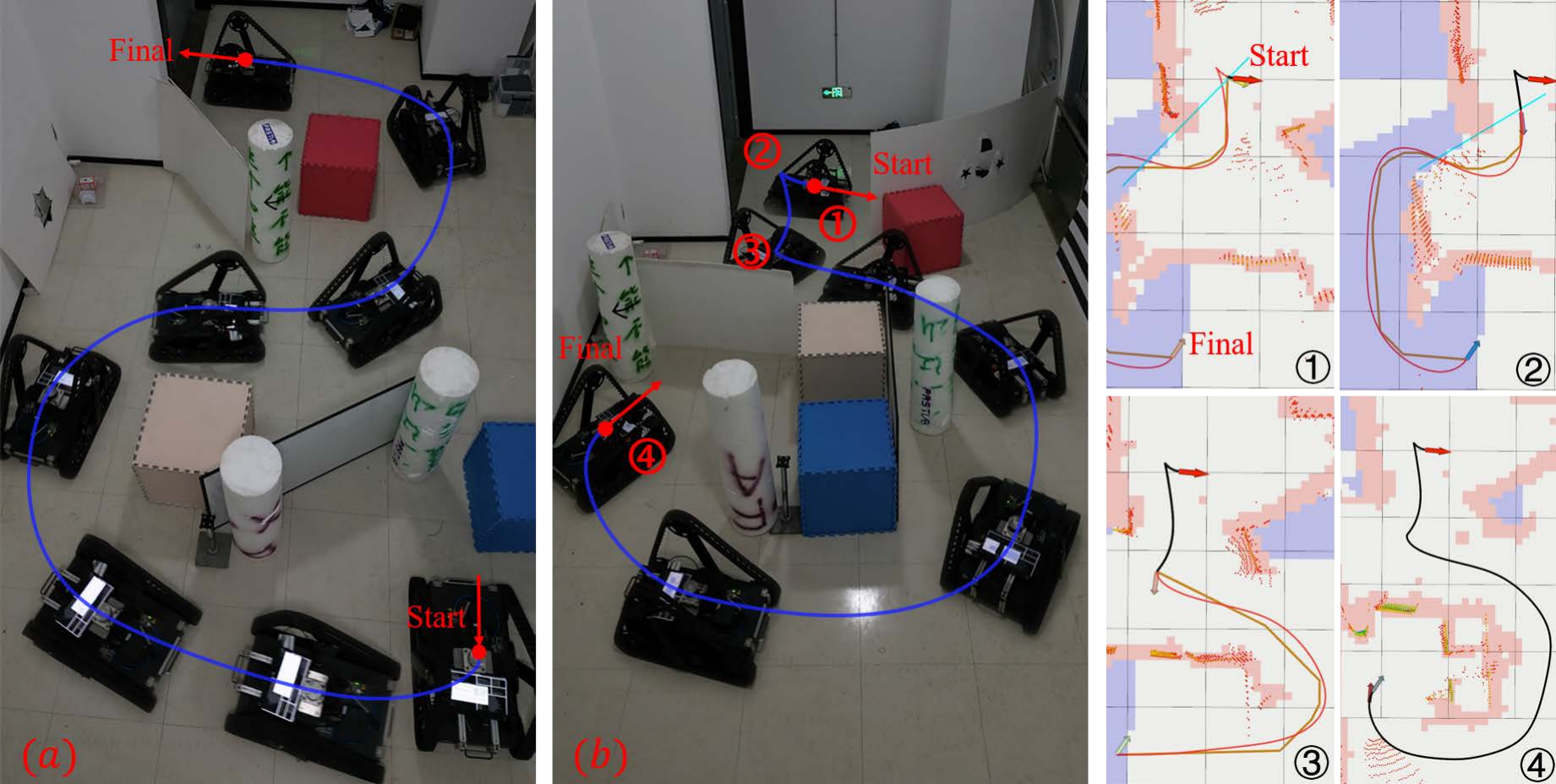}
    \caption{Planning results for the TDD robot. (a) Planning results in the complex environment. (b) Results of trajectory topology change due to occlusion.  \textcircled{1}-\textcircled{4} are parts of the replanning results of (b). Orange lines are the results of JPS, red lines are the current optimized trajectory, and black lines represent trajectories that have already been passed.}
    \label{fig:track_robot_result}
    \vspace{-0.65cm}
\end{figure}

We employ the CubeTrack \cite{10802301} as the TDD platform, which is controlled by the rotational speed of the two tracks.
We use FAST-LIO for localization and limit the perception range of the Lidar to $7\text{m}$ from the robot. 
We limit the maximum rotational speed of both tracks to $V_{max} = 0.5 \text{m/s}$ and replan at a rate of $10\text{Hz}$.
% Thanks to the additional degrees of freedom from the track arm, the planning difficulty increases. 
The rotating arms change the mass distribution of the robot and therefore affect the ICRs, which need to be estimated online.
Given the heavy mass of the robot, its linear and angular accelerations are inherently low, which necessitates a sufficiently smooth trajectory to minimize tracking errors and ensure safe navigation through narrow spaces. 
As shown in Fig.\ref{fig:track_robot_result}(a), the robot can safely reach the final position in the narrow environment.
To account for the occlusion caused by obstacles and rotating arms, the robot needs to perform additional maneuvers to explore the accessible areas.
As shown in Fig.\ref{fig:track_robot_result}(b), at position \textcircled{1}, the optimized trajectory chooses to go backwards to leave the confined space even though JPS does not specify backward obstacle avoidance. 
At position \textcircled{2}, the optimistic rule chooses to pass from the left side since no obstacle on the left side is observed. 
However, at position \textcircled{3}, the robot observes the wall and replans to leave the wrong direction in the confined space. 
Eventually, the robot reaches the final position safely.

\section{Conclusion} \label{sec:conclusion}
In this paper, we propose an efficient planning and control system for the differential drive robot class. 
We introduce a novel MS trajectory, which ensures the continuity and linearity of the higher-order kinematics of the trajectory representation while reducing computational complexity. 
We present the optimization problem and methods for handling corresponding constraints, including kinematic constraints and safety constraints specific to DD platforms. 
We provide trajectory preprocessing, parameter estimation, and an NMPC-based controller to enhance our planner and control system. 
Simulation results demonstrate the high computational efficiency of our method while ensuring the quality of the trajectory. 
In real-world experiments, we validated the efficiency and robustness of our system across various platforms.

For future work, we will further refine the planner to account for the effects of sideslip. 
Additionally, we aim to enhance the system's capabilities, including risk-awareness in replanning and the ability to handle challenging terrains, thereby expanding the potential applications of our system. 

\vspace{-0.16cm}
\bibliography{references}

\vspace{-15 mm} 
    
\begin{IEEEbiography}[{\includegraphics[width=0.9in,height=1.15in,clip,keepaspectratio]{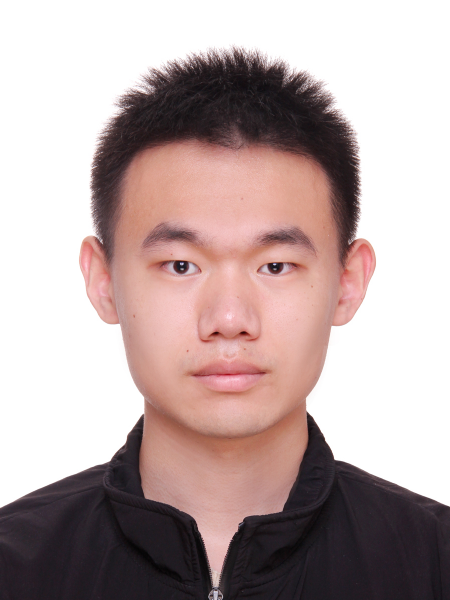}}]
    {Mengke Zhang}
    received the B.Eng. degree in Automation from Zhejiang University, Hangzhou, China, in 2021. He is currently pursuing the Ph.D. degree at the Fast Lab, Zhejiang University. His research interests include motion planning and uneven terrains.
\end{IEEEbiography}

\vspace{-15 mm} 

\begin{IEEEbiography}[{\includegraphics[width=1in,height=1.15in,clip,keepaspectratio]{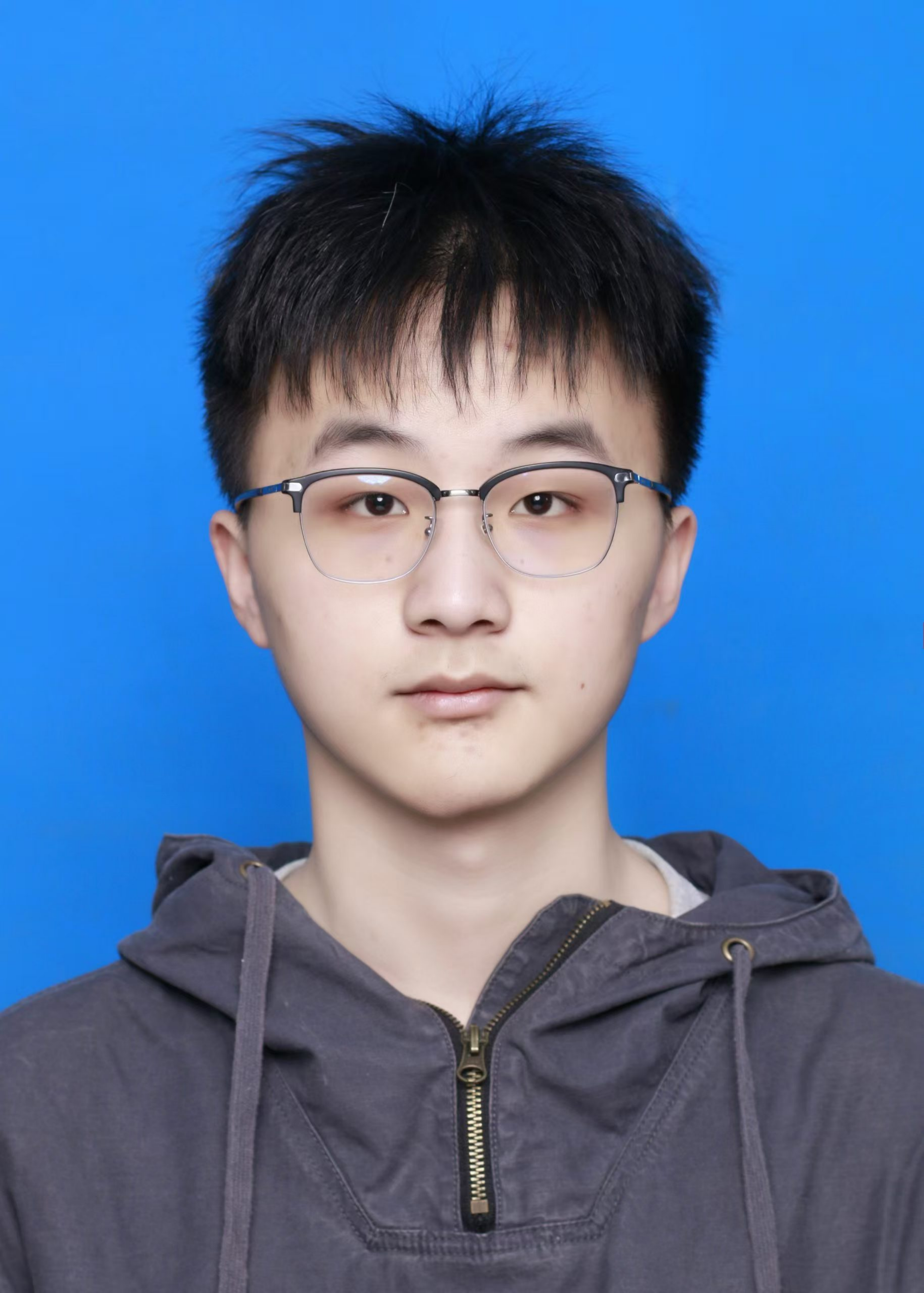}}]
    {Nanhe Chen}
    received the B.Eng. degree in robotics engineering from Zhejiang University, Hangzhou, China, in 2023. He is currently pursuing the M.S. degree in control science and engineering in the Fast Lab at Zhejiang University. His research interests include motion planning and multi-robot systems.
\end{IEEEbiography}

% \vspace{-15 mm} 

\begin{IEEEbiography}[{\includegraphics[width=1in,height=1.15in,clip,keepaspectratio]{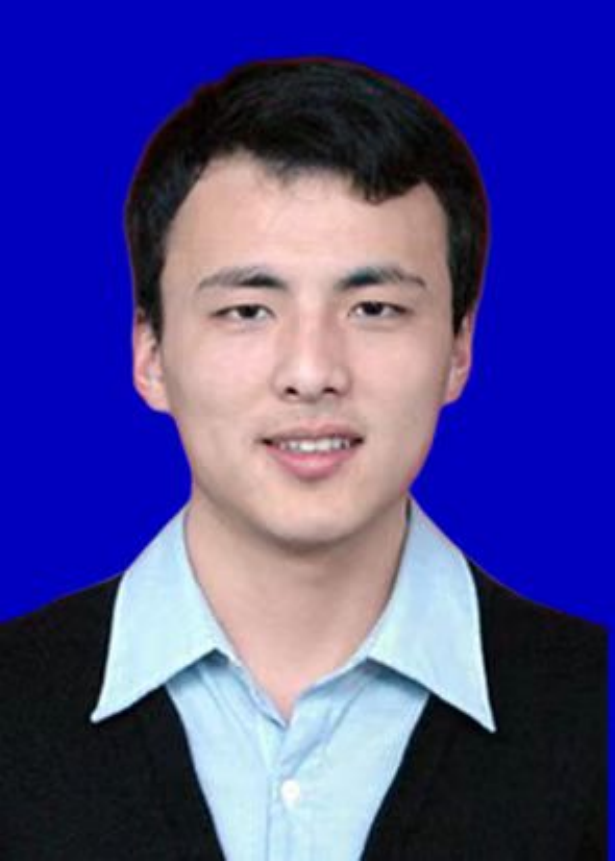}}]
    {Hu Wang}
    received the B.Eng. degree in Electrical Engineering and Automation from Huazhong University of Science and Technology. 
    He is currently working for China Tobacco Zhejiang Industrial Co., Ltd. 
    His research interests include control theory and equipment energy saving. 
\end{IEEEbiography}

\vspace{-15 mm} 

\begin{IEEEbiography}[{\includegraphics[width=1in,height=1.15in,clip,keepaspectratio]{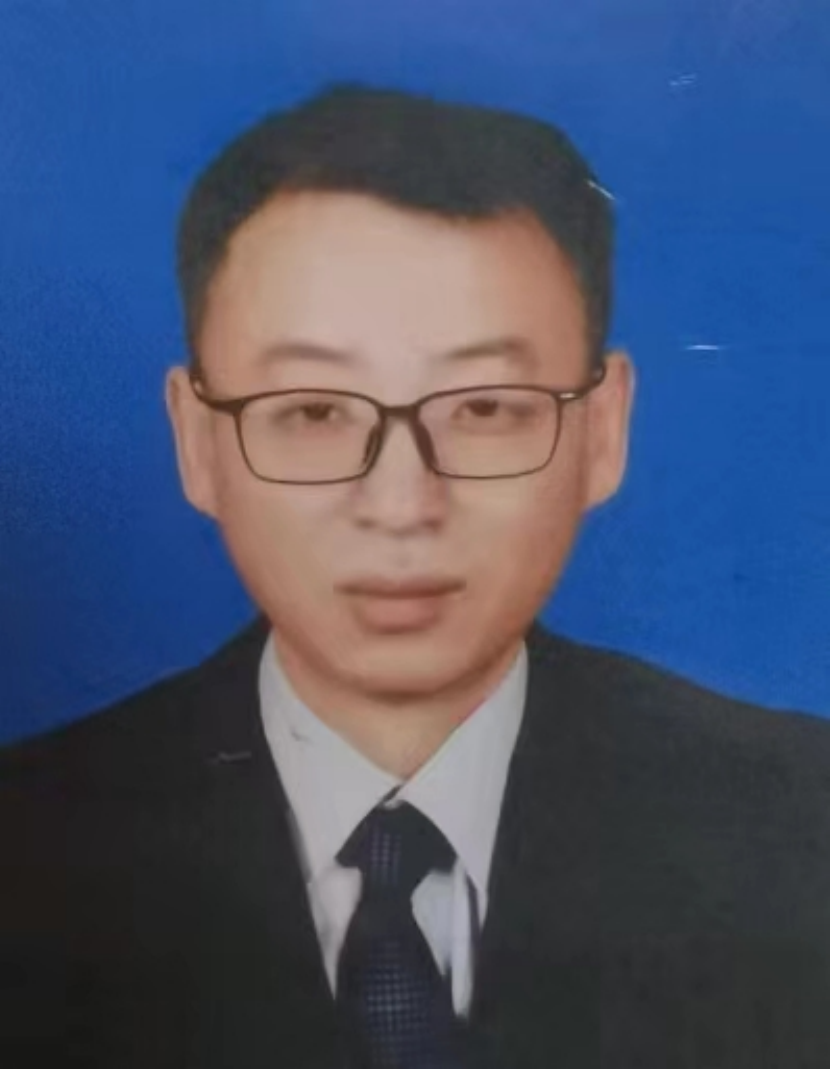}}]
    {Jianxiong Qiu} received the M.Eng. degree in Engineering. 
    He is currently working for China Tobacco Zhejiang Industrial Co., Ltd. 
    His research interests include equipment management, control research, and green energy-saving.
\end{IEEEbiography}

\vspace{-15 mm} 

\begin{IEEEbiography}[{\includegraphics[width=1in,height=1.15in,clip,keepaspectratio]{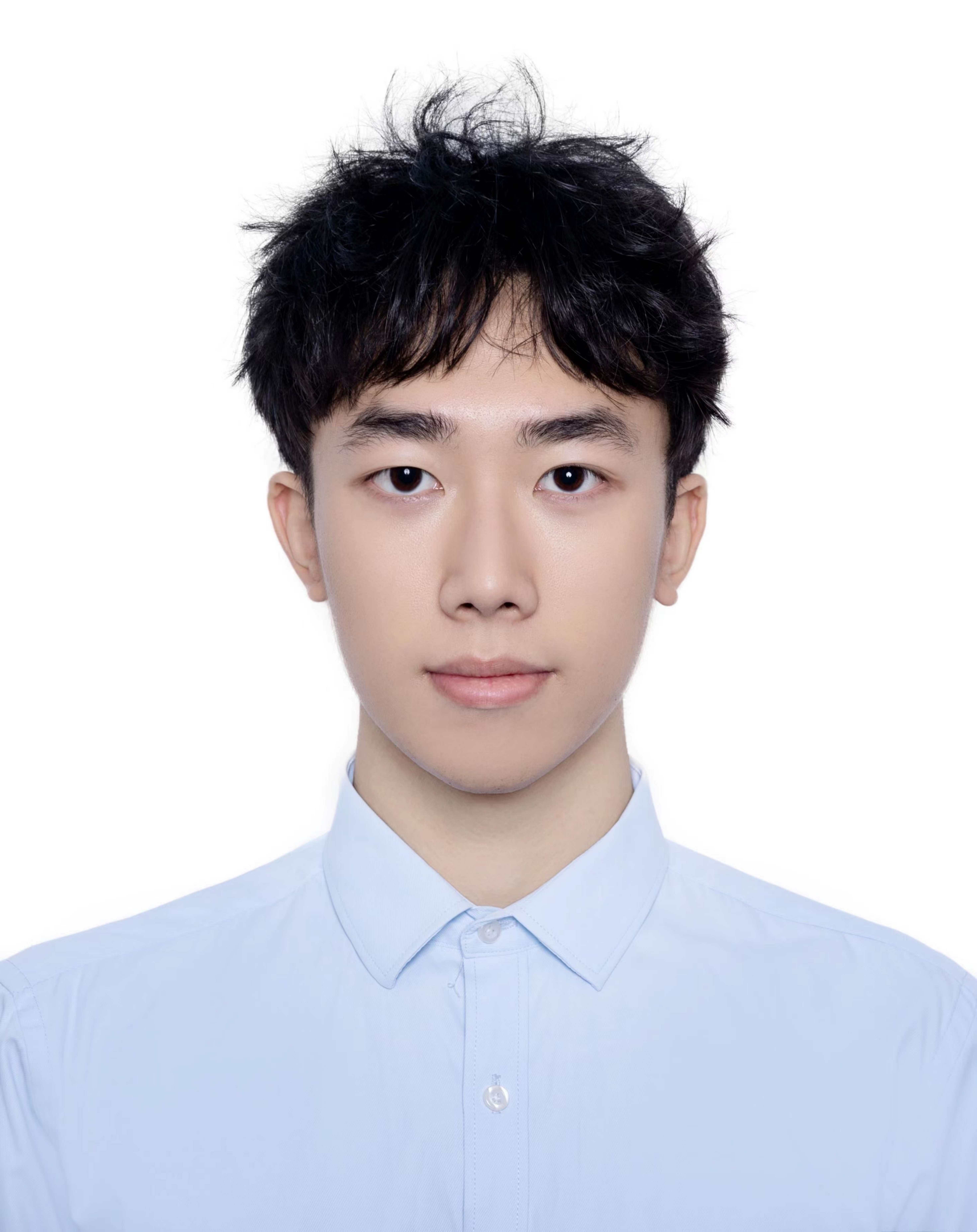}}]
    {Zhichao Han} 
    received the B.Eng. degree in Automation from Zhejiang University, Hangzhou, China, in 2021. Currently, he is actively pursuing a Ph.D. degree in the Fast Lab at Zhejiang University, where he is under the supervision of Prof. Fei Gao. His research interests include motion planning and robot learning.
\end{IEEEbiography}

\vspace{-15 mm} 

\begin{IEEEbiography}[{\includegraphics[width=1in,height=1.15in,clip,keepaspectratio]{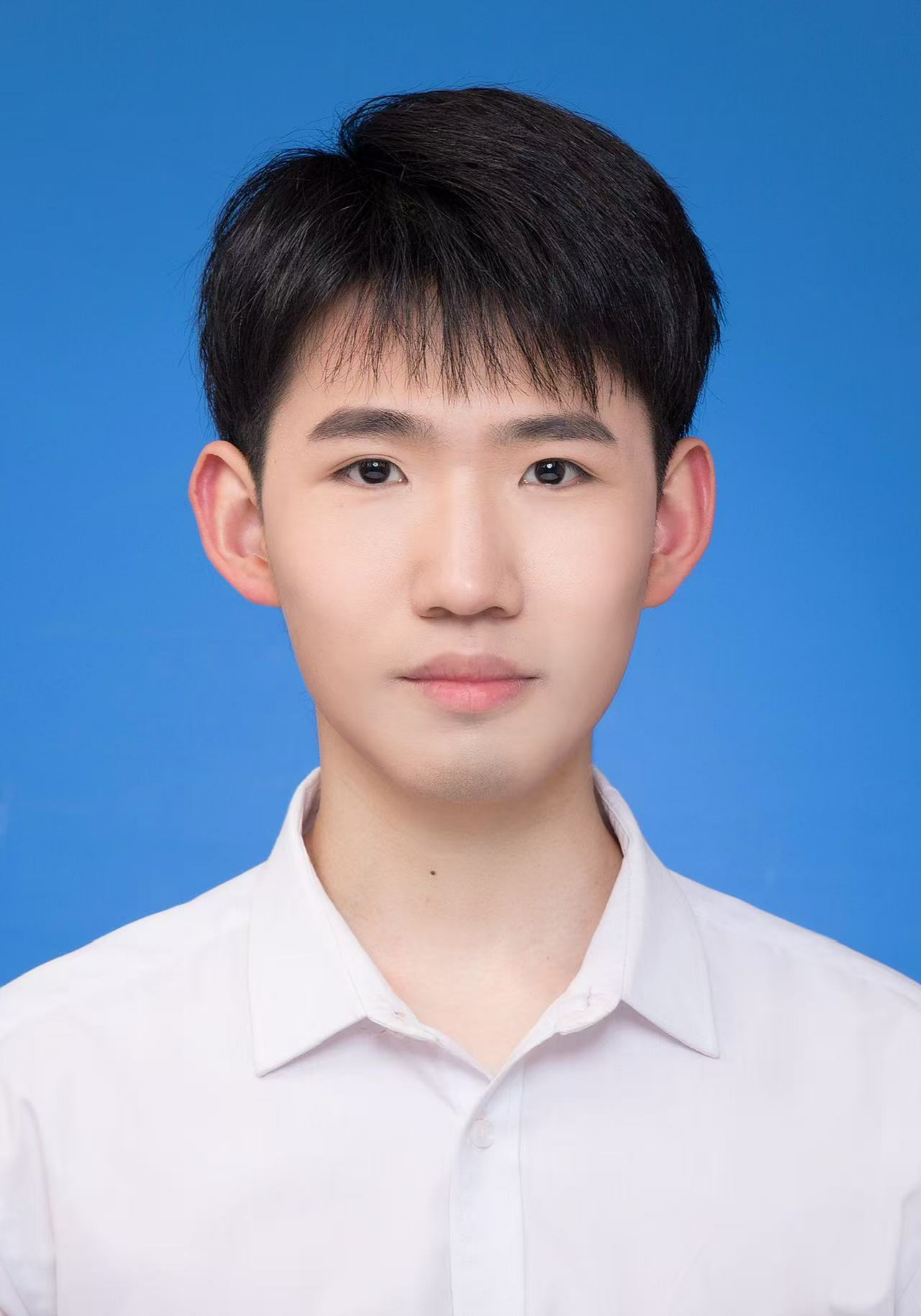}}]
    {Qiuyu Ren} 
    received the B.Eng. degree in Automation from the College of Electrical Engineering at Zhejiang University, China, in 2021. He is currently pursuing his Ph.D. in Electronic Information at the College of Control Science and Engineering, Zhejiang University. His research interests include UAV tracking and Model Predictive Control (MPC).
\end{IEEEbiography}

\vspace{-15 mm} 

\begin{IEEEbiography}[{\includegraphics[width=1in,height=1.15in,clip,keepaspectratio]{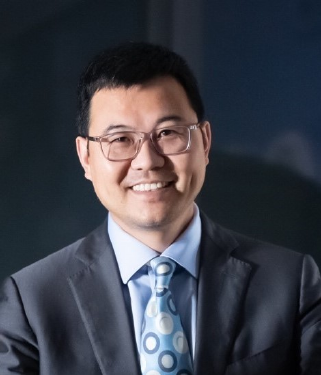}}]
    {Chao Xu}
    received his Ph.D. in Mechanical Engineering from Lehigh University in 2010. He is currently Associate Dean and Professor at the College of Control Science and Engineering, Zhejiang University. He serves as the inaugural Dean of ZJU Huzhou Institute, as well as plays the role of the Managing Editor for IET Cyber-Systems \& Robotics. His research expertise is Flying Robotics and Control-theoretic Learning. Prof. Xu has published over 100 papers in international journals, including Science Robotics (Cover Paper), Nature Machine Intelligence (Cover Paper), etc. Prof. Xu will join the organization committee of the IROS-2025 in Hangzhou.
\end{IEEEbiography}

\vspace{-13 mm} 

\begin{IEEEbiography}[{\includegraphics[width=1in,height=1.15in,clip,keepaspectratio]{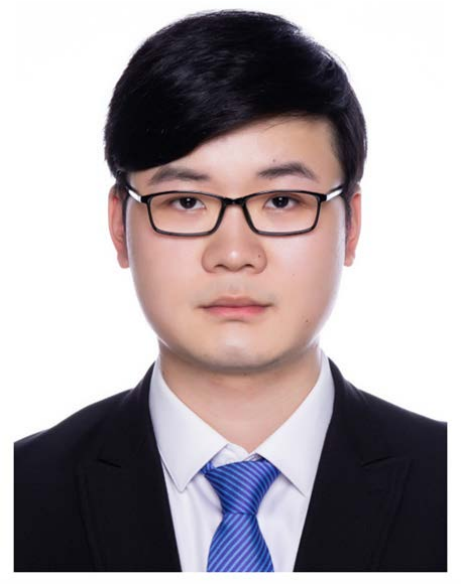}}]
    {Fei Gao}
    received the Ph.D. degree in electronic and computer engineering from the Hong Kong University of Science and Technology, Hong Kong, in 2019. He is currently a tenured associate professor at the Department of Control Science and Engineering, Zhejiang University, where he leads the Flying Autonomous Robotics (FAR) group affiliated with the Field Autonomous System and Computing (FAST) Laboratory. His research interests include aerial robots, autonomous navigation, motion planning, optimization, and localization and mapping.
\end{IEEEbiography}

\vspace{-13 mm} 

\begin{IEEEbiography}[{\includegraphics[width=1in,height=1.15in,clip,keepaspectratio]{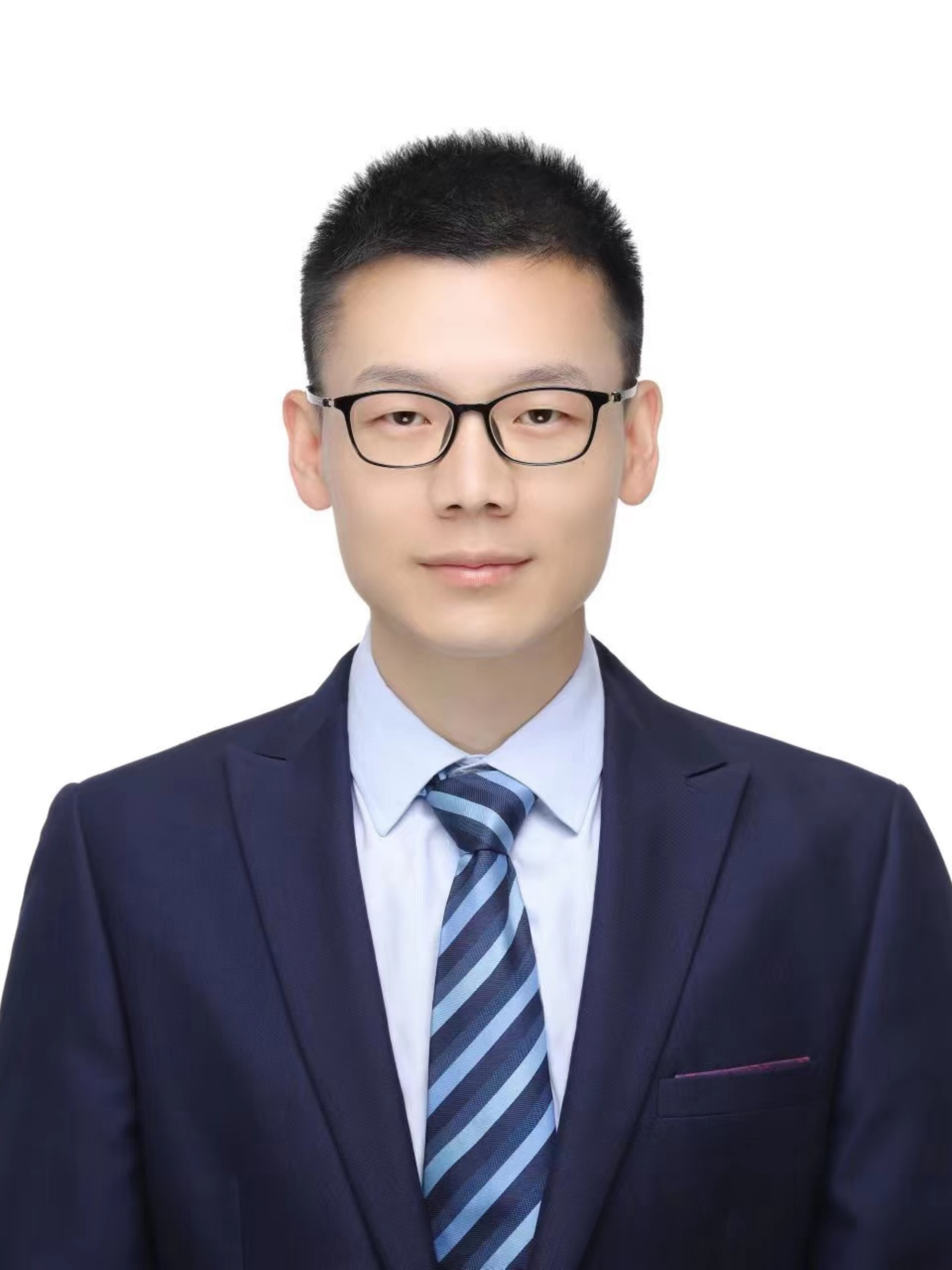}}]
    {Yanjun Cao}
    received his Ph.D. degree in computer and software engineering from the University of Montreal, Polytechnique Montreal, Canada, in 2020. He is currently an associate researcher at the Huzhou Institute of Zhejiang University, as a PI in the Center of Swarm Navigation. He leads the Field Intelligent Robotics Engineering (FIRE) group of the Field Autonomous System and Computing Lab (FAST Lab). His research focuses on key challenges in multi-robot systems, such as collaborative localization, autonomous navigation, perception and communication. 
\end{IEEEbiography}

\end{document}